\PassOptionsToPackage{table}{xcolor}
\documentclass{bmvc2k}

\title{RevalExo: A Functional Daily-Activity Benchmark for Inertial and Visual Locomotion Mode Recognition in Older Adults and Clinical Cohorts}

\addauthor{Diwas Lamsal}{diwas.lamsal@kuleuven.be}{*1}
\addauthor{Juha Carlon}{juha.carlon@kuleuven.be}{*1}
\addauthor{Reinhard Claeys}{reinhard.claeys@vub.be}{*2}
\addauthor{Maxim Yudayev}{maxim.yudayev@kuleuven.be}{1}
\addauthor{Louis Flynn}{louis.flynn@vub.be}{2}
\addauthor{Tom Verstraten}{tom.verstraten@vub.be}{2}
\addauthor{David Beckw\'ee}{david.beckwee@vub.be}{2}
\addauthor{Eva Swinnen}{eva.swinnen@vub.be}{2}
\addauthor{Mihai B\^ace}{mihai.bace@kuleuven.be}{$\dagger$1}
\addauthor{Bart Vanrumste}{bart.vanrumste@kuleuven.be}{$\dagger$1}
\addauthor{Benjamin Filtjens}{b.filtjens@tudelft.nl}{$\dagger$3}

\addinstitution{
 KU Leuven\\
 Leuven, Belgium
}
\addinstitution{
 Vrije Universiteit Brussel\\
 Brussels, Belgium
}
\addinstitution{
 Delft University of Technology\\
 Delft, The Netherlands
}
\runninghead{Lamsal et al.}{RevalExo}

\usepackage{booktabs}
\usepackage{amssymb}
\usepackage[table]{xcolor}
\definecolor{modHeader}{HTML}{E8E8E8}  %
\usepackage{diagbox}
\usepackage{graphicx}
\usepackage{array}
\usepackage{adjustbox}
\usepackage{multirow}
\usepackage{algorithm}
\usepackage{algpseudocode}

\begin{document}

\maketitle

\makeatletter
\BMVA@blfootnote{$^{*}$Equal contribution. $^{\dagger}$Equal senior contribution.}
\makeatother

\begin{abstract}
Assistive devices for people with mobility impairments, such as powered exoskeletons, rely on accurate locomotion mode recognition to adapt control strategies and provide appropriate assistance during daily activities. However, public benchmarks are typically collected from healthy adults, lack temporally precise labels necessary for detecting mode transitions, or focus on a limited set of tasks. To support development and evaluation under realistic clinical constraints and daily mobility demands, we introduce RevalExo, a functional daily-activity benchmark for inertial and visual locomotion mode recognition. RevalExo is built around a standardized, clinically and ecologically validated daily-activity protocol reflecting the cumulative everyday mobility demands in ageing and clinical populations. The benchmark includes 27 participants across three cohorts: older adults without mobility impairments, stroke survivors, and older adults with probable sarcopenia. The full cohort was recorded with lower-body IMUs, while synchronized egocentric video was collected for a clinically feasible subset of 13 participants. RevalExo provides 10.1 hours of frame-level annotations across 11 locomotion modes, including 5.1 hours of paired inertial--visual recordings. We benchmark three challenges: unimodal and multimodal locomotion mode recognition across multiple horizons, cross-population generalization from older adults without mobility impairments to clinical cohorts, and vision-guided knowledge transfer to IMU-only models. Results confirm consistent gains from fusing inertial and visual inputs but reveal a substantial gap between general recognition ($\sim$93\% F1) and recognition during transitions ($\sim$68\% F1), alongside persistent challenges in cross-population generalization and cross-modal transfer. We release RevalExo to stimulate further research on these open challenges.\footnote{RevalExo is available at: \url{https://revalexo.github.io/}}

\end{abstract}

\section{Introduction}
\label{sec:intro}

Stroke remains a leading cause of long-term disability worldwide, with nearly 12 million cases annually and 94 million people living with its effects~\cite{Feigin2025WSO}. Sarcopenia, the age-related loss of muscle mass, affects an estimated 10--27\% of older adults~\cite{PetermannRocha2022SarcopeniaPrev}, increasing risk of falls and injuries~\cite{Beaudart2025GLIS}. For these populations, restoring independent mobility is a central rehabilitation goal, driving interest in wearable assistive devices such as lower-limb exoskeletons~\cite{Cha2025WearableRobots}.

Effective assistance from such devices requires accurate \emph{locomotion mode recognition}: identifying the type of locomotion the user is performing to adapt support accordingly~\cite{10682545} and ideally predicting upcoming changes in advance so the device can adjust in time~\cite{10633818,10682545}. Inertial Measurement Units (IMUs) are widely used for their portability and low cost~\cite{Samala2024ProsthesisDataset}, but are blind to the environment: they can capture how a person moves, but not what lies ahead~\cite{9896501}. Egocentric vision fills this gap with a continuous first-person view of the terrain~\cite{chen2024enhancing, 9896501, 9098903, 9437301krausz}, but is vulnerable to occlusion and lighting variation and often requires high-capacity models that are difficult to deploy on wearable hardware~\cite{zhang2019sensor, Tschiedel2020RelyingOnMore}. Because inertial signals capture user movement dynamics, combining them with vision can reduce reliance on high-capacity visual feature extractors while improving overall recognition performance, driving interest in multimodal inertial--visual systems~\cite{10682545, chen2024enhancing, Sharma2023, 9437301krausz, 10160419tsepa}.

Clinical evaluation of locomotion mode recognition systems requires benchmarks that reflect the daily realities of mobility-impaired individuals~\cite{Hu2018ENABL3S}. Public locomotion datasets, however, typically cover a narrow set of tasks such as level walking or stair/slope navigation~\cite{10605888sftik,9896501} and are collected mostly from healthy adults, while the few existing clinical locomotion datasets are limited by their small cohort sizes, IMU-only sensing, or simplified protocols such as treadmill walking~\cite{Samala2024ProsthesisDataset,Ahkami2025}. Public multimodal resources remain scarce: where synchronized egocentric video has been released, it covers unrestricted public-space walking without frame-level locomotion mode annotations~\cite{Sharma2023}. No public benchmark therefore captures impaired or altered gait in clinical populations across diverse activities of daily living (ADLs) with temporally precise labels necessary for assistive control.

Three further deployment challenges shape what such a benchmark must address. First, \emph{transition phases}, the brief periods when the user switches between modes, are harder to recognize than steady-state locomotion because they are rare and therefore underrepresented in training data, individually variable, and ambiguous: sensor readings during a transition do not cleanly match either the preceding or upcoming mode~\cite{MarcosMazon2022,DalPreteTransition}. Yet these are precisely the moments at which assistive control must react. Second, standard evaluation assumes training and test data come from the same distribution, but collecting at scale from clinical populations is rarely feasible, motivating evaluation of \emph{cross-population generalization}: whether models trained on older adults without mobility impairments remain robust for clinical cohorts whose gait kinematics can differ substantially. Third, even when richer modalities such as egocentric video can be collected, additional visual equipment can be invasive or burdensome for patient populations, motivating \emph{vision-guided knowledge transfer}, where models leverage vision during training but operate on IMU-only inputs at deployment~\cite{KDSurvey,CrossModalSurvey}.

We introduce \textbf{RevalExo}, a benchmark dataset for inertial and visual locomotion mode recognition under realistic clinical constraints. RevalExo is built around FATIG'AGE, a standardized, clinically and ecologically validated daily-activity protocol co-created with older adults, stroke survivors, and clinicians to capture the sequenced and cumulative demands of everyday mobility~\cite{PERCEPT,PERCEPT2,FATIGAGE}. Data collection was shaped by clinical feasibility: all 27 participants, spanning older adults without mobility impairments (HC; also referred to as non-impaired older adults; $N{=}7$), stroke survivors (ST; $N{=}10$), and older adults with probable sarcopenia (SR; also referred to as sarcopenic older adults; $N{=}10$), were recorded with seven lower-body IMUs, while synchronized egocentric video was collected for the 13 participants who safely tolerated the visual equipment. The resulting dataset provides 10.1 hours of frame-level annotations across 11 locomotion modes, including 5.1 hours of paired inertial--visual recordings. We benchmark three challenges: (1) locomotion mode recognition with inertial, visual, and multimodal inputs across multiple prediction horizons (0--1\,s), with separate evaluation during transitions; (2) cross-population generalization from older adults without mobility impairments to the clinical cohorts; and (3) vision-guided knowledge transfer, where egocentric video is leveraged during training to improve IMU-only inference on clinical cohorts. Results confirm consistent gains from inertial--visual fusion but reveal a substantial gap between overall and transition F1 ($\sim$93\% vs.\ $\sim$68\%), alongside persistent challenges in cross-population generalization and cross-modal transfer.

\section{Related Work}

\begin{table*}[t]
\begin{center}
\small
\setlength{\tabcolsep}{4.2pt}
\renewcommand{\arraystretch}{1.08}
\begin{tabular}{|l|c|c|c|c|c|c|}
\hline
\textbf{Dataset} & \textbf{Subjects (N)} & \textbf{Clinical} & \textbf{IMU} & \textbf{Ego-Vision} & \textbf{Hours} & \textbf{\# Classes} \\
\hline
Hu et al.~\cite{Hu2018ENABL3S} & 10 & $\times$ & $\checkmark$ (5) & $\times$ & $\sim$6.5 & 7 \\
Camargo et al.~\cite{Camargo2021} & 22 & $\times$ & $\checkmark$ (4) & $\times$ & 19.9 & 6 \\
Laschowski et al.~\cite{Laschowski2020exonet} & 1 & $\times$ & $\times$ & $\checkmark$ & $>$52 & 12 \\
Sharma et al.~\cite{Sharma2023} & 23$^{\dagger}$ & $\times$ & $\checkmark$ (17) & $\checkmark$ & 24.2 (11.8)$^{\dagger}$ & 6 \\
Zhao et al.~\cite{10605888sftik} & 10 & $\times$ & $\checkmark$ (3) & $\checkmark$ & $\sim$5 & 5 \\
Samala et al.~\cite{Samala2024ProsthesisDataset} & 30 & $\checkmark$ & $\checkmark$ (4) & $\times$ & $\sim$0.5 & 1 \\
Ahkami et al.~\cite{Ahkami2025} & 5 & $\checkmark$ & $\checkmark$ (3) & $\times$ & $\sim$5 & 5 \\
\hline
\textbf{RevalExo (ours)} & \textbf{27 (13)}$^{*}$ & $\checkmark$ & \textbf{$\checkmark$ (7)} & $\checkmark$ & \textbf{10.1 (5.1)}$^{*}$ & \textbf{11} \\
\hline
\end{tabular}
\end{center}
\caption{Comparison of RevalExo with existing human locomotion datasets by subject count, clinical coverage, presence of sensor modalities, dataset size in number of hours, and number of annotated classes. IMU numbers indicate sensor count. $^{\dagger}$Sharma et al.\ report 24.2\,h of IMU recordings, including 11.8\,h of synchronized egocentric video from 23 subjects. $^{*}$RevalExo includes lower-body IMU recordings from 27 participants (10.1\,h), with synchronized egocentric video for a 13-participant subset (5.1\,h).}

\label{tab:dataset_comparison}
\end{table*}

\subsection{Locomotion Mode Recognition Datasets}

Public human locomotion datasets vary along several axes: cohort size, clinical coverage, sensor modalities, recording duration, and label granularity. Table~\ref{tab:dataset_comparison} compares RevalExo with some representative examples.

Early work focused on kinematics and physiology. Hu et al.~\cite{Hu2018ENABL3S} (ENABL3S) recorded lower-limb IMU and electromyography (EMG) from 10 participants during transitions between seven locomotion modes. Camargo et al.~\cite{Camargo2021} collected motion-capture and inertial data from 22 participants across six locomotion modes.

Egocentric vision has been explored for terrain-aware locomotion mode recognition~\cite{9098903,9437301krausz}. Laschowski et al.~\cite{Laschowski2020exonet} (ExoNet) provide a large collection of egocentric images of walking environments from one participant without synchronized inertial data. Recent resources combine inertial and visual sensing: Sharma et al.~\cite{Sharma2023} provide full-body IMU and egocentric video from 23 participants, but their labels cover only session-level scenarios (e.g., ``6-floor staircase'') rather than frame-level mode boundaries; Zhao et al.~\cite{10605888sftik} provide chest-mounted RGB-D and lower-body IMU data from 10 participants across five locomotion modes.

Beyond these task-specific datasets, large-scale multimodal corpora offer a different kind of resource. Nymeria~\cite{Nymeria}, for example, provides over 300 hours of egocentric video, full-body IMU, and motion-language descriptions from 264 participants. While such resources lack the frame-level locomotion mode labels required for direct evaluation, they offer a basis for pretraining transferable representations~\cite{Le2024TransferLearning,CAREPD} that can be fine-tuned on task-specific benchmarks, thereby complementing rather than replacing clinically grounded resources.

Despite this progress, most public locomotion datasets feature healthy participants, with very little clinical coverage. Although clinical datasets often support rehabilitation outcome assessment or gait analysis~\cite{StrokeRehab,CAREPD,GAITGEN,Filtjens2022}, they typically lack standardized benchmarks and temporally precise labels for locomotion mode recognition. Datasets including mobility-impaired populations, such as amputees, often use simplified protocols (e.g., treadmill walking)~\cite{Samala2024ProsthesisDataset} or have small cohort sizes~\cite{Ahkami2025}. This gap matters because impaired gait can differ substantially from healthy gait~\cite{Feigin2025WSO,Beaudart2025GLIS}. RevalExo addresses this by combining the FATIG'AGE protocol with inertial sensing and synchronized egocentric video where clinically feasible, yielding frame-level annotations across 11 locomotion modes in non-impaired older adults and clinical cohorts. To our knowledge, RevalExo is the only public resource combining paired modalities, frame-level labels, and clinical-cohort coverage for method development and evaluation under realistic clinical conditions. 

\subsection{Methods for Locomotion Mode Recognition}
Hybrid Convolutional Neural Network (CNN) and Long Short-Term Memory (LSTM) architectures are popular for IMU-based locomotion mode recognition, where convolutional layers extract local features and LSTM layers capture temporal dynamics~\cite{9447716,MarcosMazon2022,Tang2024}. DeepConvLSTM (DCL) and its variants~\cite{Ordez2016DeepConvLSTM,BockDeepConvLSTM} have demonstrated strong performance in activity recognition, making them popular baselines. Recent studies continue to benchmark against DCL variants when evaluating vision- or fusion-based methods~\cite{BockWEAR,Zhang2024IMUVideoMAE}.

For vision, real-time inference on wearable hardware favors lightweight architectures. Prior work has typically used compact CNNs for single-frame classification, with models like MobileNet~\cite{mobilenet} and ResNet~\cite{resnet} offering good speed-accuracy trade-offs~\cite{Laschowski2022,9098903}. Beyond single-frame methods, video architectures better capture short-term temporal dynamics preceding mode changes. Efficient networks such as MoViNet~\cite{movinet} and X3D~\cite{x3d} provide lightweight feature extraction suitable for edge deployment~\cite{10719798usemovinet}. Larger transformer-based methods like MViT~\cite{MViT} prioritize accuracy over efficiency and serve as strong upper bounds.

When both IMU and egocentric vision are available, a standard strategy is to fuse learned modality representations. Feature-level fusion via concatenation (or simple averaging at the logit level) remains a strong, widely used baseline for locomotion mode recognition~\cite{CONTRERASCRUZ2023108656,10160419tsepa,11122304Shin}. More recently, fusion mechanisms such as adaptive instance normalization (AdaIN)~\cite{10160419tsepa} and sandwich fusion transformers~\cite{10605888sftik} have also been explored to improve multimodal integration while maintaining computational efficiency. Self-supervised multimodal pretraining offers an alternative path: IMU-Video-MAE~\cite{Zhang2024IMUVideoMAE}, for example, learns joint IMU--video representations via masked autoencoding for downstream use.

In our experiments, we benchmark methods across this spectrum, from lightweight baselines for real-time deployment to higher-capacity methods prioritizing accuracy. We further compare which modalities and methods best transfer to unseen clinical cohorts. 

\subsection{Vision-Guided Knowledge Transfer}
Although multimodal sensing can improve locomotion mode recognition, deploying the full inertial--visual sensor suite is not always practical. Egocentric video can raise privacy concerns and increase power and compute demands~\cite{Chen2025COMODO}, and wearable camera systems may be burdensome or stigmatizing for users. This motivates settings where rich modalities are available during training, but models must operate with a reduced sensor set at deployment (e.g., IMU-only)~\cite{CrossModalSurvey}. Beyond these deployment constraints, IMU-based recognition is also known to degrade substantially under population shift~\cite{Cai2025GeneralizableHAR}, making vision a particularly useful source of training-time supervision for IMU-only models targeting clinical cohorts.

Two complementary strategies leverage training-time vision to support IMU-only deployment. The first is knowledge distillation, where a teacher trained on video (or multimodal inputs) guides an IMU-only student. Following Gou et al.~\cite{KDSurvey}, distillation objectives transfer response-based knowledge via softened logits~\cite{HintonKD,NKD} or feature-based knowledge via intermediate representations~\cite{FitNets}, with cross-modal extensions addressing the gap when teacher and student use different modalities~\cite{CRD}. The second strategy, cross-modal representation alignment, pretrains an IMU encoder by aligning its embeddings with paired video in a shared latent space via contrastive objectives~\cite{CLIP-pmlr-v139-radford21a}, as in IMU2CLIP~\cite{moon-etal-2023-imu2clip}. Recent evidence suggests such cross-modal pretraining improves out-of-distribution generalization, including to clinical populations~\cite{Cheshmi2025OODHAR}. RevalExo provides a benchmark for this setup: paired inertial--visual recordings are available for part of the dataset, while some clinical cohorts are IMU-only due to clinical feasibility constraints. We therefore evaluate whether vision-guided training improves IMU-only inference on clinical cohorts lacking visual recordings.

\section{Dataset}
\label{sec:dataset}

\begin{figure*}[t]
\begin{center}
\begin{tabular}{cc}
\bmvaHangBox{\includegraphics[width=0.61\textwidth]{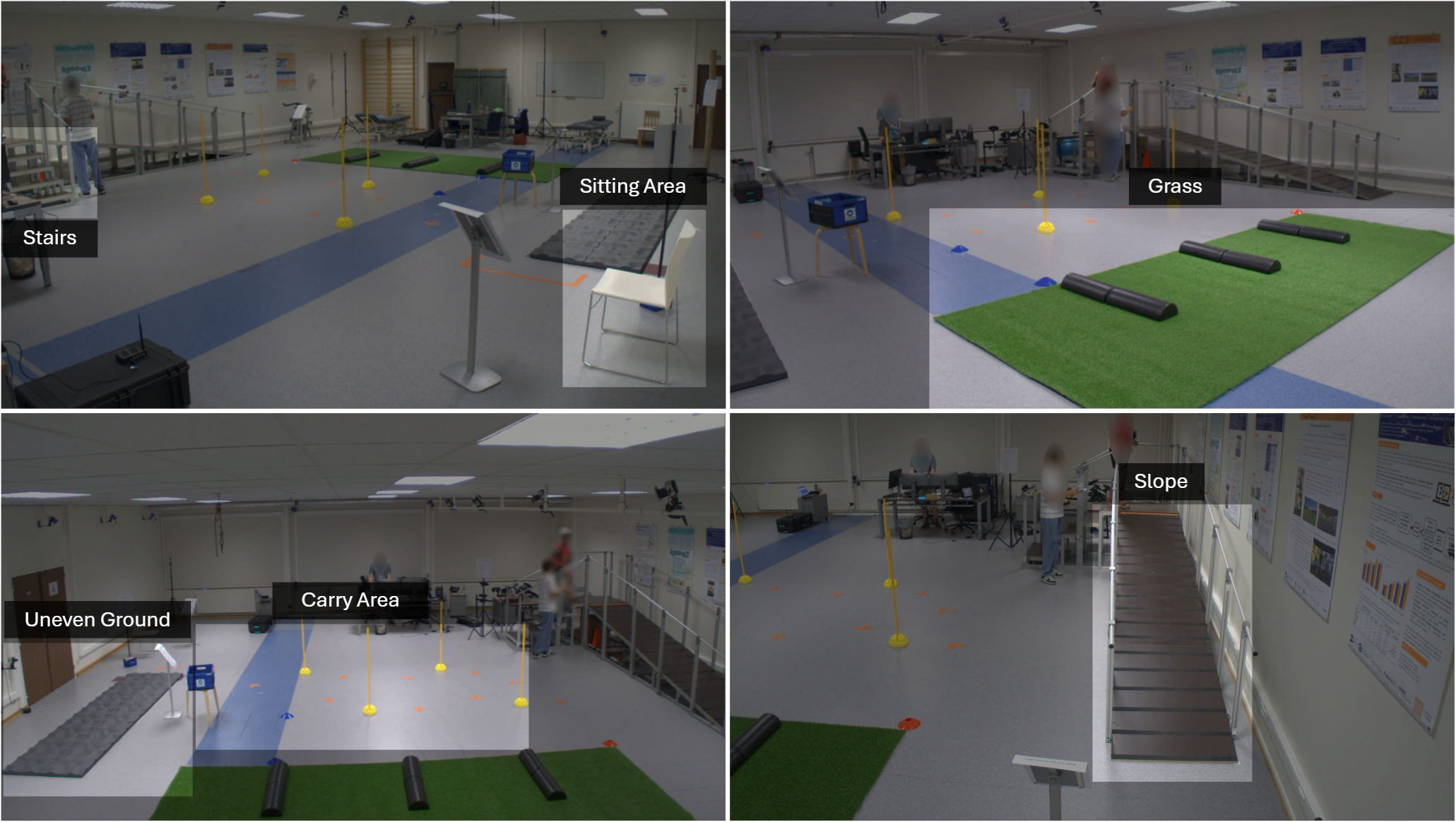}} &
\bmvaHangBox{\includegraphics[width=0.32\textwidth]{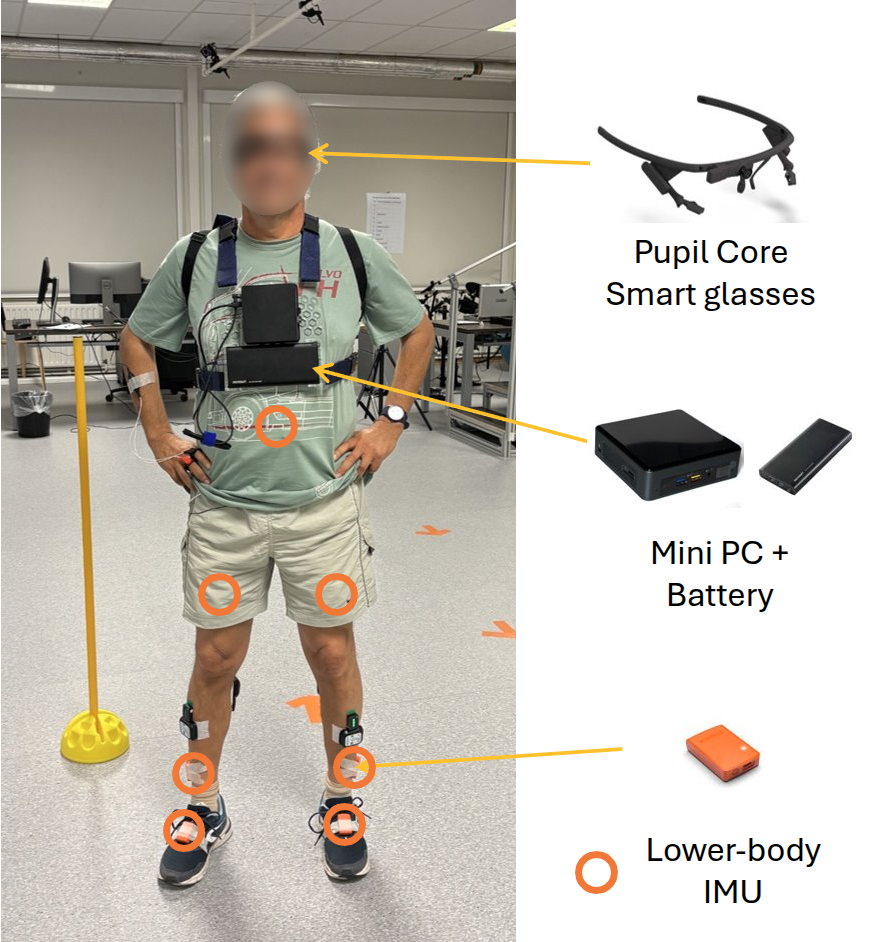}} \\
(a) & (b)
\end{tabular}
\end{center}
\caption{Overview of the RevalExo acquisition setup. (a) Data acquisition environment with highlighted task areas; see Table~\ref{tab:tasks} for the full list of locomotion modes. (b) Sensor setup with Pupil Core smart glasses~\cite{pupillabsPupilCore} for egocentric video, seven Xsens IMUs~\cite{movellaXsensAwinda} for lower-body inertial data, and a battery-powered mini PC for on-body data logging.}
\label{fig:combined-setup}
\end{figure*}

This section describes the dataset acquisition environment and sensor setup (Figure~\ref{fig:combined-setup}), cohorts, protocol, annotation process, and dataset statistics.

\subsection{Acquisition and Sensor Setup}
Data were collected in an indoor environment featuring diverse terrain and obstacles representing daily mobility challenges~\cite{PERCEPT,PERCEPT2,FATIGAGE}. Figure~\ref{fig:combined-setup} shows the setup overview (a) and sensor configuration (b). Acquisition was managed using HERMES~\cite{HERMES}, an open-source framework for synchronized multimodal sensing. It synchronized lower-body IMU, egocentric video, and external camera streams at the time of recording, enabling frame-level alignment across modalities.

\textbf{IMUs.}
Lower-body motion data were captured for all participants using seven Xsens Awinda IMUs (Movella, Netherlands)~\cite{movellaXsensAwinda}, placed according to the standard Xsens lower-body configuration. Calibration and data processing were performed using Xsens MVN Analyze software. IMU data were sampled at 60\,Hz and include raw accelerometer, gyroscope, and magnetometer measurements, as well as processed lower-body joint kinematics.

\textbf{Egocentric video.}
For the multimodal subset, egocentric video was captured using Pupil Core smart glasses (Pupil Labs, Germany)~\cite{pupillabsPupilCore} at 30 frames per second with a resolution of $1280\times720$ pixels. Video was recorded through an Intel mini PC worn in a chest pack.

\textbf{External cameras.}
To support frame-level annotation, four external Basler cameras recorded third-person views at $1920\times1080$ pixels and 30\,fps. These views were used to identify gait events, task boundaries, and ambiguous transitions during annotation.

\subsection{Participants}

The study was approved by the Medical Ethics Committee of Vrije Universiteit Brussel (VUB) [EC-2024-203, BUN 1432024000164]. All participants were briefed on the experimental protocol and signed an informed consent form. Sensor data are fully anonymized, retaining no personally identifiable information. Researchers occasionally visible in the egocentric video recordings provided separate consent for the public release of these recordings.

Participants were recruited into three cohorts with specific eligibility criteria. Older adults without mobility impairments were included if $\geq$65 years old and excluded if they experienced mobility restrictions according to the International Classification of Functioning, Disability and Health~\cite{WHO2001ICF}. Stroke survivors were included if $\geq$18 years old, $\geq$6 months post-stroke, and had a Functional Ambulation Category (FAC) of 3 (requiring verbal supervision or the presence of another person while walking) or 4 (independent walking limited to level surfaces); they were excluded for severe speech or memory problems preventing understanding of instructions, or for full dependence on a walker or mobility scooter. \emph{Probable sarcopenia} denotes reduced muscle strength without confirmation of reduced muscle quantity or quality~\cite{CruzJentoft2019}. Older adults with probable sarcopenia were included if $\geq$65 years old and demonstrated reduced muscle strength, defined as a five-times sit-to-stand time $>$15 seconds and low handgrip strength ($<$71 kPa for men, $<$41 kPa for women)~\cite{CruzJentoft2019,BSGGSarcopeniaGuidelines}; those dependent on a walker or mobility scooter, or unable to understand instructions, were excluded.

\subsection{Protocol}
\label{sec:protocol}
RevalExo follows FATIG'AGE, a standardized protocol designed to capture the sequenced and cumulative mobility demands of everyday life in ageing and clinical populations~\cite{PERCEPT,PERCEPT2,FATIGAGE}. In RevalExo, this protocol defines the task sequence, repeated-trial structure, stopping criteria, and safety supervision used during data collection. Before each session, all equipment was prepared and verified. Participants first completed a screening questionnaire and physical assessments to verify eligibility against the cohort-specific inclusion criteria. Sensors were then placed and calibrated. Given the inclusion of clinical cohorts with a high fall risk, a researcher continuously supervised each trial in close proximity to provide immediate assistance in the event of loss of balance.

Each trial began in sitting and included sit-to-stand and stand-to-sit repetitions, followed by stair up, ramp down, grass walking, uneven-ground walking, carry, ramp up, and stair down, before returning to the chair area and ending in sitting (Figure~\ref{fig:combined-setup}(a)). Level-ground walking connected all standing and walking tasks. Participants completed repeated trials for up to one hour, with sessions ending early if the participant reported fatigue or if the supervising clinical researcher judged continuation inappropriate.

The egocentric video setup added approximately 1.4\,kg of additional load through the chest-mounted mini PC and battery pack. Piloting indicated that this load posed two major concerns: a direct safety risk for stroke survivors with hemiplegic shoulder drop and reduced usable session time in participants with probable sarcopenia, who reach fatigue limits much earlier than other cohorts under the protocol (see Section~\ref{sec:dataset_statistics}). Egocentric video recording was therefore omitted for 14 participants (4 stroke survivors and all 10 older adults with probable sarcopenia), who completed the protocol with IMU-only sensing.

\subsection{Annotation}

\begin{table}[t]
\begin{center}
\footnotesize 
\begin{tabular}{|c|p{1.8cm}|l|}
\hline
 & Label & Description \\
\hline
1  & Level ground  & Walk or stand on flat surface. \\
2  & Sit to stand  & Rise from a chair. \\
3  & Stand to sit  & Sit down onto a chair. \\
4  & Sitting       & Remain seated on a chair. \\
5  & Stair up      & Ascend a flight of stairs. \\
6  & Stair down    & Descend a flight of stairs. \\
7  & Ramp up       & Walk up an incline. \\
8  & Ramp down     & Walk down an incline. \\
9  & Grass         & Walk on grass with obstacles. \\
10 & Uneven ground & Walk on an irregular surface. \\
11 & Carry         & Carry weights and traverse a short path
with turns. \\
\hline
\end{tabular}
\end{center}
\caption{The 11 locomotion modes in the RevalExo dataset.}
\label{tab:tasks}
\end{table}

We performed frame-level annotations using ELAN~\cite{ELAN}, which supports simultaneous playback of all four external camera streams. Three trained student annotators (bachelor's and master's level) labeled the data using a reference guide and live demonstrations. Each frame was assigned exactly one of the 11 locomotion modes (Table~\ref{tab:tasks}), following established practice~\cite{10633818,10682545}. Transitions are not labeled as a separate class but are defined at evaluation time as predictions whose target time falls within $\pm 0.25$\,s of a mode boundary~\cite{MarcosMazon2022,Kang2022TransitionEvaluation} (see Section~\ref{sec:task}). Boundary criteria followed prior locomotion studies~\cite{MarcosMazon2022,9447716,10682545,10633818}. For walking-based tasks (e.g., stair, ramp, grass, uneven ground), onset was the toe-off initiating movement toward the task area and offset was the last heel contact completing the task. For seated transitions, sit-to-stand onset was the first observable trunk motion (e.g., forward lean) and offset was a stable upright posture; stand-to-sit followed the inverse criteria. For the carry task, onset was the upward movement of the weight and offset was placing it down. Ambiguous cases were flagged and resolved with supervising researchers.

To quantify reliability, all three annotators independently labeled a subset of three subjects spanning the three cohorts ($\sim$10\% of the data). Frame-level agreement across all 11 classes was high, with Fleiss' $\kappa=0.944$ and mean pairwise Cohen's $\kappa=0.943$ (range 0.933--0.955). Per-category $\kappa$ exceeded 0.92 for 9 of 11 classes, with lower values for the two shortest classes: sit-to-stand (0.868) and stand-to-sit (0.832). For boundary agreement, we computed boundary F1 at multiple tolerances: 0.866 at $\pm$250\,ms and 0.957 at $\pm$500\,ms, with median start and end boundary discrepancies of 67\,ms and 100\,ms, respectively. The triple-annotated subset and computation scripts are released alongside the dataset.

\subsection{Dataset Statistics}
\label{sec:dataset_statistics}

\begin{table*}[t]
\begin{center}
\scriptsize 
\begin{tabular}{|l|c|c|c|c|c|c|}
\hline
\textbf{Cohort} & \textbf{$n$} & \textbf{Age (yrs)} & \textbf{Height (cm)} & \textbf{Weight (kg)} & \textbf{Sex (M/F)} & \textbf{Duration (h)} \\
\hline
\multicolumn{7}{|l|}{\textit{Multimodal (inertial + visual)}} \\
\hline
\quad HC & 7  & 74.9$\pm$7.2  & 168.4$\pm$9.1  & 72.9$\pm$12.4 & 3/4   & 3.1 \\
\quad ST & 6  & 56.8$\pm$6.6  & 166.3$\pm$10.1 & 78.2$\pm$23.7 & 2/4   & 2.1 \\
\hline
\multicolumn{7}{|l|}{\textit{Inertial only}} \\
\hline
\quad ST & 4  & 58.7$\pm$15.5 & 177.4$\pm$11.5 & 68.9$\pm$10.6 & 3/1   & 1.9 \\
\quad SR & 10 & 84.3$\pm$4.4  & 166.7$\pm$12.4 & 76.7$\pm$9.6  & 4/6   & 3.0 \\
\hline
Overall  & 27 & 71.9$\pm$14.2 & 168.6$\pm$11.0 & 74.9$\pm$14.1 & 12/15 & 10.1 \\
\hline
\end{tabular}
\end{center}
\caption{Participant demographics. Age, height, and weight reported as mean $\pm$ SD; duration is the total recording time per cohort. HC: non-impaired older adults; ST: stroke survivors; SR: sarcopenic older adults.}
\label{tab:demographics}
\end{table*}

\begin{figure}[t]
    \centering
    \includegraphics[width=0.95\columnwidth]{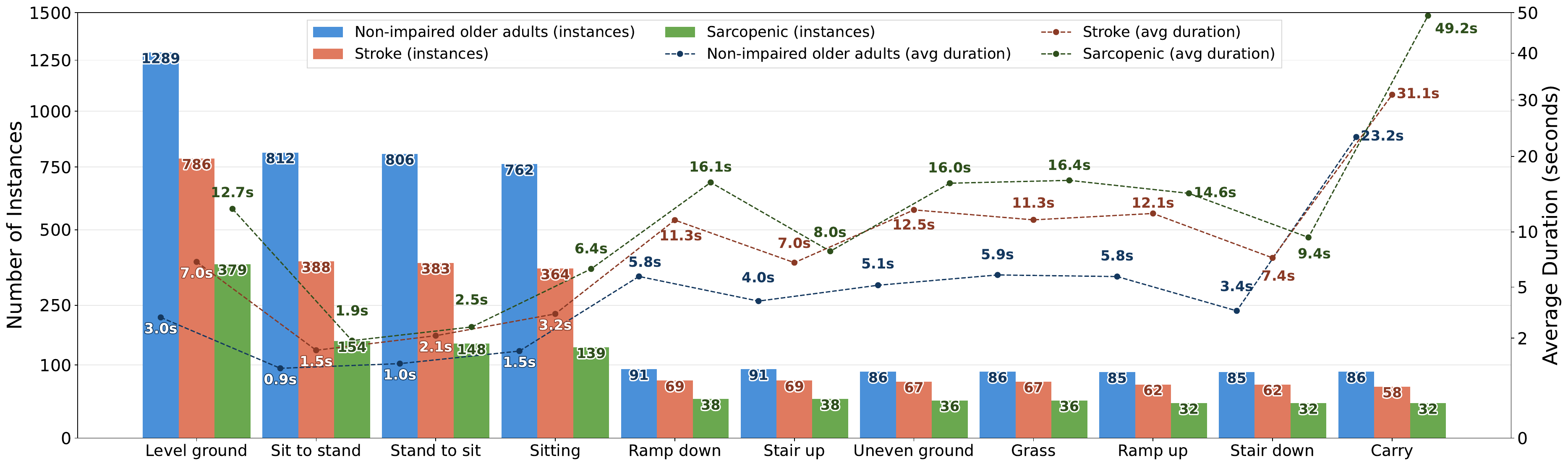}
    \caption{Per-cohort number (left axis) and average duration (right axis) of instances per task. An instance is a single, uninterrupted segment of a task.}
    \label{fig:task-statistics}
\end{figure}

Table~\ref{tab:demographics} summarizes the dataset demographics. RevalExo contains 10.1 hours of annotated data from 27 participants across three cohorts. Synchronized egocentric video is available for the multimodal subset (7 non-impaired older adults and 6 stroke survivors, $\sim$5.1 hours); the remaining 14 participants contribute inertial-only recordings.

Figure~\ref{fig:task-statistics} reports per-cohort instance counts (bars) and average segment durations (lines) across the 11 locomotion modes. Level-ground walking is the most frequent class in every cohort and accounts for the largest share of time. Terrain and obstacle tasks (stair, ramp, grass, uneven ground) are less frequent but longer. For almost every class, instance counts decrease and average durations increase from non-impaired older adults to stroke survivors to sarcopenic older adults. Carry has the longest segments, averaging $23.2$\,s for non-impaired older adults and $49.2$\,s for sarcopenic older adults, who also contribute fewer instances.

\section{Experimental Setup}
\label{sec:setup}

RevalExo's primary target is locomotion mode recognition. This section describes the task formulation, evaluation protocol, and methods used across the three benchmark tasks.

\subsection{Task and Evaluation Setup}
\label{sec:task}

\begin{figure}[t]
    \centering
    \includegraphics[width=0.75\columnwidth]{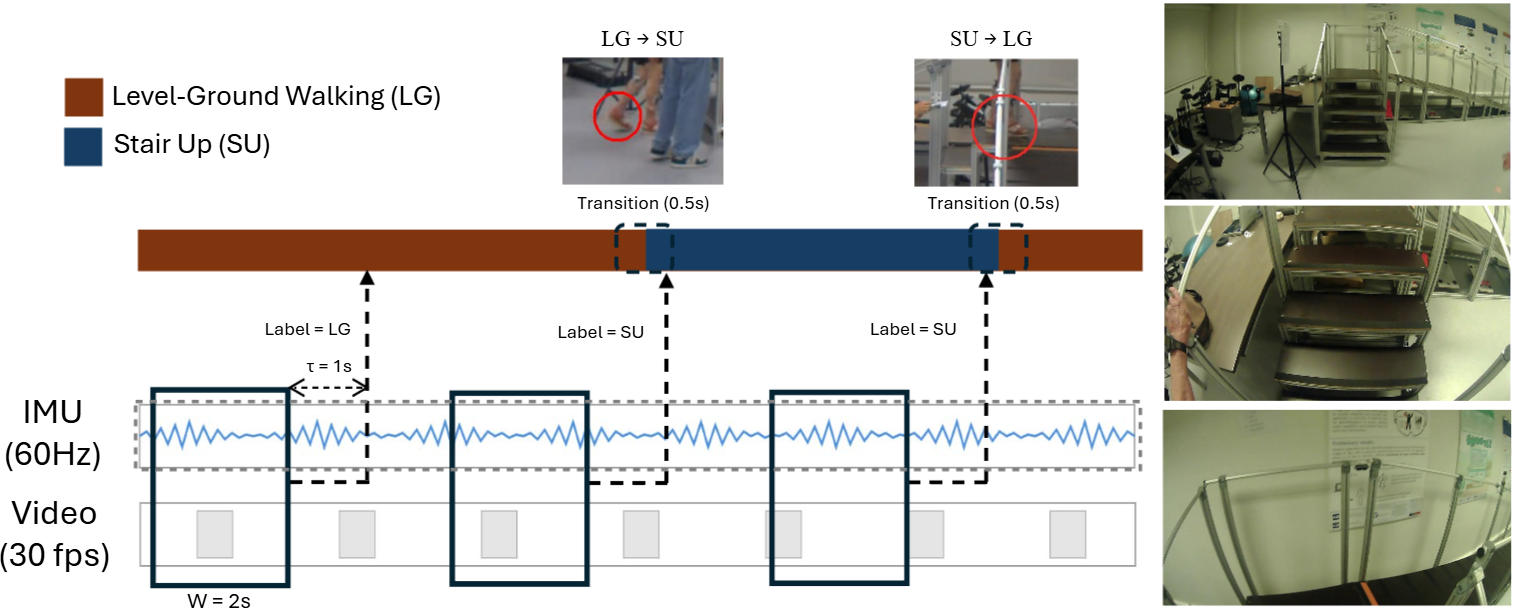}
    \caption{The locomotion mode recognition task (left) alongside an example sequence seen through the egocentric view (right). Models use inertial, visual, or both modalities to predict the locomotion mode. The first toe-off during the level-ground$\rightarrow$stair-up transition marks the start of stair up, and the final heel contact marks the end. Predictions whose target time falls within $\pm 0.25$\,s of these boundaries are counted as transition predictions.}
    \label{fig:task_definition}
\end{figure}

We formulate locomotion mode recognition as a multi-horizon classification problem. Given an input $\mathbf{x}_t$ ending at time $t$, the model predicts the locomotion mode $y_{t+\tau}$ at $t + \tau$, with $\tau \in \{0, 0.1, 0.2, 0.3, 0.5, 1.0\}$\,s spanning the typical range of lead times reported in prior work~\cite{10633818,10380790,Carvalho2025} (Figure~\ref{fig:task_definition} shows an example at $\tau{=}1$\,s). Evaluating across this range allows researchers to select appropriate models based on their application requirements. At $\tau{=}0$ the task reduces to \emph{current-state recognition}; for $\tau>0$ it becomes \emph{predictive recognition}, with the mode at $t+\tau$ used as the target even when it differs from the current mode at $t$.

The input $\mathbf{x}_t$ is a 2\,s temporal window for the inertial and video modalities, or a single frame for the image modality. The 2\,s duration matches the 1--2\,s range that best trades off latency and accuracy for inertial-based recognition~\cite{WindowSize} (verified empirically in Appendix~B) and is close to the temporal windows used to train video architectures such as X3D~\cite{x3d} and MViT~\cite{MViT}. To ensure a matching number of training and test examples across modalities, the same window duration is applied to both temporal modalities.

\textbf{Metrics.} We report macro F1 averaged across test subjects. Following standard practice~\cite{10633818,10682545,Kang2022TransitionEvaluation}, we also report performance on \emph{transition windows}, whose target time $t+\tau$ falls within $\pm 0.25$\,s of a locomotion mode change~\cite{MarcosMazon2022}. We report results at $\tau\in\{0,0.5\}$\,s; full results across all six horizons are included in Appendix~E.

\textbf{Data Preparation.} Long inertial--visual recordings were split into short, synchronized clips to avoid decoding long sequences in the data loader~\cite{gong2023cavmae,Zhang2024IMUVideoMAE}. During training, multiple windows are sampled per clip; during evaluation, a deterministic sliding window with a stride of 0.25\,s is applied. All models operate on single frames or 2\,s windows with no access to trial-level context or session ordering. Inter-task gaps contain level-ground walking segments that usually exceed the 2\,s input window, making the fixed FATIG'AGE task sequence largely invisible to the benchmarked models. Validation uses an 80/20 clip-level split of the training-subject clips; the held-out test subjects are evaluated at the end of training.

\textbf{Benchmark Tasks.} We evaluate three tasks. \textbf{(1)~Locomotion mode recognition:} using leave-one-subject-out cross-validation (LOSO-CV) on the 13 subjects of the multimodal subset (7 non-impaired older adults and 6 stroke survivors). \textbf{(2)~Cross-population generalization:} training on non-impaired older adults ($N{=}7$), testing unimodal and multimodal models on stroke survivors with paired data ($N{=}6$), and additionally evaluating inertial-only DCL on the full stroke and sarcopenic cohorts ($N{=}10$ each). \textbf{(3)~Vision-guided knowledge transfer:} leveraging video at training time to improve an inertial-only student that is trained on non-impaired older adults and then evaluated on the clinical cohorts. Separately, we assess \textbf{deployment feasibility} of the benchmarked methods on wearable devices by measuring end-to-end inference throughput on a Jetson Orin Nano (8\,GB, 15\,W).

\subsection{Methods}
\label{sec:methods}
All methods are implemented in PyTorch and trained for 50 epochs using original hyperparameters where available. Each model's decoder is replaced by a shared multi-horizon classification head operating on the same feature representation. Class-balanced sampling is used across all experiments to oversample underrepresented classes during training. Full implementation details are in the supplementary codebase.

\textbf{Unimodal baselines.} For the inertial modality, we use DCL following the configuration from Bock et al.~\cite{BockWEAR}. For the visual modality, we use ResNet~\cite{resnet} and MobileNet-v3~\cite{mobilenet} for single-frame classification, and CNN-based X3D~\cite{x3d} and transformer-based MViT-Base (16$\times$4)~\cite{MViT} to capture spatiotemporal information from video clips.

\textbf{Multimodal fusion.} We fuse the unimodal encoders via feature-level concatenation or averaging, allowing each encoder to retain its original architecture while combining complementary kinematic and visual cues at the representation level. We additionally evaluate KIFNet~\cite{10160419tsepa} (MLP + MobileOne~\cite{10204436mobileone}), SFTIK~\cite{10605888sftik} (SFTIK-RGB variant), and IMU-Video-MAE~\cite{Zhang2024IMUVideoMAE} (using two feet and two upper-leg IMUs to match their 4-IMU configuration).

\textbf{Vision-guided transfer.} We compare two strategies: (1)~contrastive pretraining (CP) following IMU2CLIP~\cite{moon-etal-2023-imu2clip}, which aligns inertial and X3D-XS video representations in a shared latent space before training the inertial encoder; and (2)~knowledge distillation, where a frozen ResNet-50+DCL teacher guides an inertial-only student (DCL with acc+gyro). For distillation, we evaluate vanilla KD ($T{=}4$, $\alpha{=}0.5$)~\cite{HintonKD,Cho2019Efficacy}, FitNet-style feature distillation~\cite{FitNets}, CRD ($\tau{=}0.07$, memory bank $N{=}16384$)~\cite{CRD}, and NKD ($\gamma{=}1.5$)~\cite{NKD}. All methods are compared against a baseline trained without vision guidance.

\section{Results}
\label{sec:results}

\subsection{Locomotion Mode Recognition}

\begin{table*}[t]
\centering
\begin{minipage}[t]{0.49\textwidth}
\vspace*{0pt}
\centering
\scriptsize
\setlength{\tabcolsep}{1.5pt}
\renewcommand{\arraystretch}{1.06}
\begin{tabular}{lcccc}
\toprule
\textbf{Method} & \multicolumn{2}{c}{$\boldsymbol{\tau{=}0.0}$\,s} & \multicolumn{2}{c}{$\boldsymbol{\tau{=}0.5}$\,s} \\
\cmidrule(lr){2-3} \cmidrule(lr){4-5}
 & Overall & Trans. & Overall & Trans. \\
\midrule
\rowcolor{modHeader}\multicolumn{5}{l}{\textit{\textbf{Inertial}}} \\
DCL (a)        & 78.6$\pm$8.6  & 50.6$\pm$9.8  & 70.9$\pm$9.2  & 46.4$\pm$8.1 \\
DCL (a/g)      & 80.8$\pm$8.0  & 53.4$\pm$10.1 & 74.2$\pm$8.1  & 47.7$\pm$8.3 \\
\midrule
\rowcolor{modHeader}\multicolumn{5}{l}{\textit{\textbf{Image}}} \\
MNV3-S         & 80.2$\pm$2.6  & 52.1$\pm$3.3  & 81.5$\pm$2.3  & 50.5$\pm$3.7 \\
R18            & 82.8$\pm$2.1  & 53.8$\pm$3.0  & 83.2$\pm$1.7  & 52.2$\pm$3.2 \\
R50            & 82.8$\pm$1.8  & 54.2$\pm$3.1  & 83.6$\pm$2.0  & 52.5$\pm$3.3 \\
\midrule
\rowcolor{modHeader}\multicolumn{5}{l}{\textit{\textbf{Video}}} \\
X3D-XS         & 86.1$\pm$3.0  & 47.8$\pm$6.9  & 82.4$\pm$4.6  & 40.0$\pm$7.9 \\
MViT           & 90.9$\pm$2.2  & 62.9$\pm$3.6  & 87.8$\pm$2.2  & 58.0$\pm$3.5 \\
\midrule
\rowcolor{modHeader}\multicolumn{5}{l}{\textit{\textbf{Inertial + Image}}} \\
KIFNet-Style   & 84.7$\pm$2.5  & 51.5$\pm$7.0  & 80.8$\pm$2.8  & 47.8$\pm$6.7 \\
KIFNet (A)     & 87.5$\pm$2.7  & 55.2$\pm$7.9  & 83.7$\pm$2.1  & 49.3$\pm$6.4 \\
KIFNet (C)     & 87.0$\pm$3.0  & 55.5$\pm$6.1  & 83.6$\pm$2.0  & 50.8$\pm$5.8 \\
MNV3-DCL (A)   & 91.4$\pm$1.7  & 65.1$\pm$6.1  & 88.1$\pm$2.2  & 57.8$\pm$5.6 \\
MNV3-DCL (C)   & 91.3$\pm$1.6  & 65.8$\pm$4.0  & 87.9$\pm$2.4  & 59.4$\pm$4.7 \\
R18-DCL (A)    & 92.3$\pm$1.7  & 65.9$\pm$5.3  & 88.5$\pm$2.0  & 58.7$\pm$4.2 \\
R18-DCL (C)    & 92.1$\pm$1.7  & 66.2$\pm$5.0  & 88.6$\pm$2.1  & 59.3$\pm$5.5 \\
\midrule
\rowcolor{modHeader}\multicolumn{5}{l}{\textit{\textbf{Inertial + Video}}} \\
SFTIK          & 89.2$\pm$3.5  & 59.8$\pm$4.8  & 86.9$\pm$2.9  & 58.0$\pm$4.4 \\
IMU-Video-MAE  & 90.3$\pm$2.0  & 61.8$\pm$4.8  & 88.0$\pm$2.4  & 57.5$\pm$5.8 \\
X3D-XS-DCL (A) & 90.1$\pm$3.8  & 56.2$\pm$7.7  & 83.2$\pm$5.6  & 41.3$\pm$9.0 \\
X3D-XS-DCL (C) & 91.1$\pm$2.9  & 60.2$\pm$6.9  & 85.7$\pm$4.0  & 47.3$\pm$7.1 \\
MViT-DCL (A)   & 92.7$\pm$1.7  & 67.6$\pm$4.6  & 88.8$\pm$2.5  & 61.0$\pm$5.2 \\
MViT-DCL (C)   & \textbf{92.8$\pm$2.0} & \textbf{68.2$\pm$5.4} & \textbf{89.6$\pm$2.3} & \textbf{61.9$\pm$4.5} \\
\bottomrule
\end{tabular}
\caption{Locomotion mode recognition at $\tau{=}0.0$\,s and $\tau{=}0.5$\,s. Mean $\pm$ SD macro F1 (\%) over the 13 multimodal subjects (LOSO-CV). \textbf{Bold}: best per column. DCL: DeepConvLSTM; MNV3-S: MobileNet-v3 Small; R18/R50: ResNet-18/-50; a/g: accelerometer/gyroscope; A/C: feature averaging/concatenation.}
\label{tab:locomotion_mode_recognition_results}
\end{minipage}\hfill
\begin{minipage}[t]{0.49\textwidth}
\vspace*{0pt}
\centering
\makeatletter\def\@captype{table}\makeatother
\scriptsize
\setlength{\tabcolsep}{2pt}
\renewcommand{\arraystretch}{1.06}
\begin{tabular}{|l|*{5}{>{\centering\arraybackslash}m{3.1em}}|}
\hline
\diagbox[width=6.7em, height=3.5em, innerleftsep=2pt, innerrightsep=2pt]{\textbf{Class}}{\textbf{Method}}
 & \rotatebox[origin=lB]{68}{DCL (a/g)}
 & \rotatebox[origin=lB]{68}{R50}
 & \rotatebox[origin=lB]{68}{MViT}
 & \rotatebox[origin=lB]{68}{R18-DCL (C)}
 & \rotatebox[origin=lB]{68}{MViT-DCL (C)} \\
\noalign{\vspace*{-0em}}
\hline
Level ground  & \underline{78.6} & 87.4 & 91.4 & 91.8 & \textbf{93.3} \\
Sit to stand  & 83.5 & \underline{51.6} & 79.4 & \textbf{85.0} & 84.1 \\
Stand to sit  & 77.1 & \underline{46.2} & 80.9 & 83.6 & \textbf{86.1} \\
Sitting       & 89.5 & \underline{82.8} & 88.7 & \textbf{90.3} & 90.0 \\
Stair up      & \underline{86.8} & 89.1 & 92.2 & 93.8 & \textbf{94.9} \\
Stair down    & \underline{83.6} & 87.1 & 94.2 & 93.9 & \textbf{95.1} \\
Ramp up       & \underline{86.3} & 92.5 & 93.3 & 94.7 & \textbf{94.8} \\
Ramp down     & \underline{84.5} & 92.3 & 92.9 & \textbf{94.4} & 93.7 \\
Grass         & \underline{76.1} & 94.4 & 95.4 & 95.8 & \textbf{96.7} \\
Uneven ground & \underline{80.3} & 93.6 & 95.6 & 95.6 & \textbf{96.6} \\
Carry         & \underline{62.8} & 93.2 & 95.8 & 93.9 & \textbf{95.9} \\
\hline
\end{tabular}
\caption{Subject-averaged per-class F1 (\%) at $\tau{=}0$\,s for the top models per modality. \textbf{Bold}: best per class. \underline{Underline}: worst per class.}
\label{tab:per_class_f1}

\vspace{2em}

\makeatletter\def\@captype{figure}\makeatother
\includegraphics[width=\linewidth]{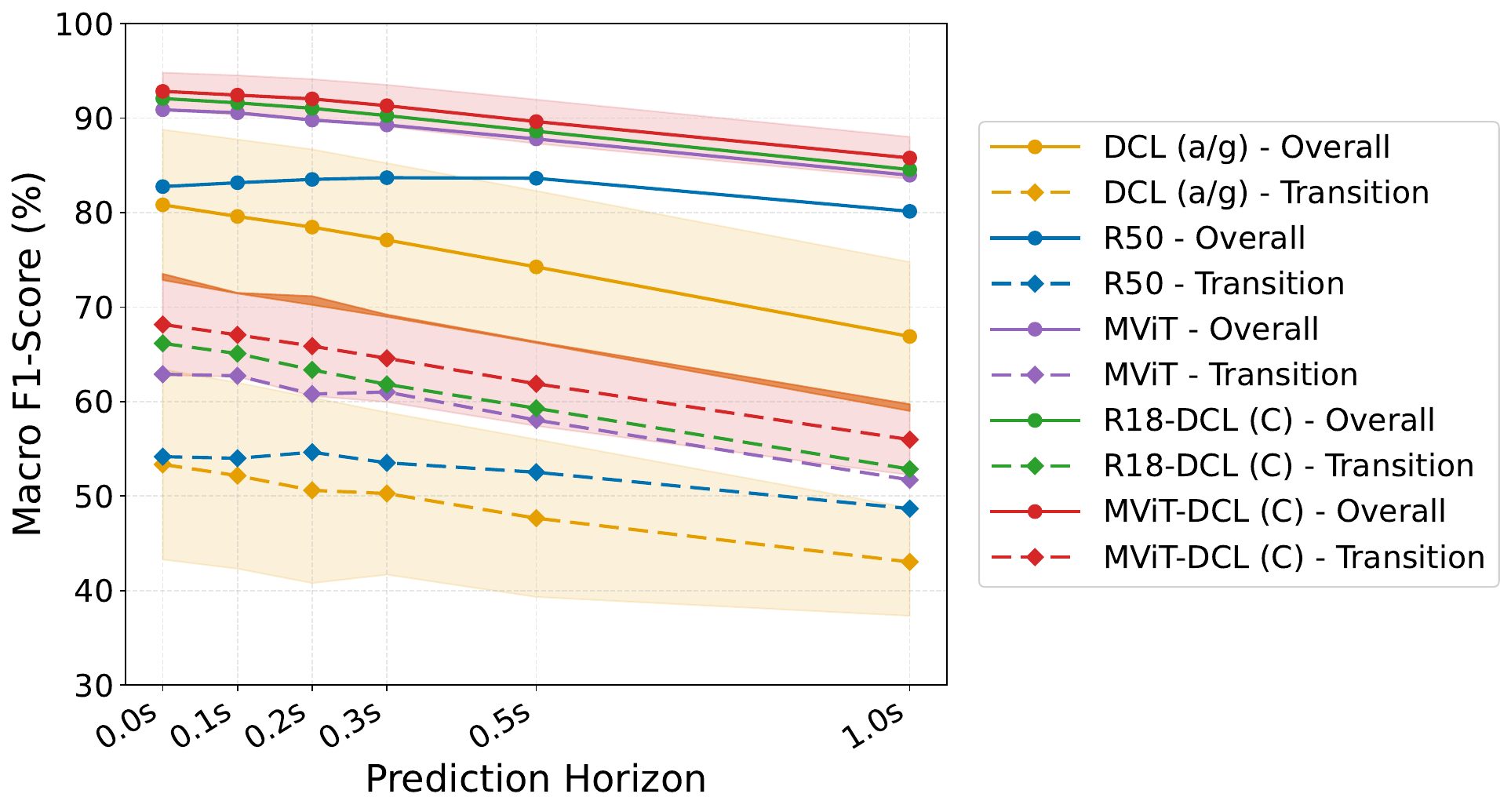}
\caption{Overall (solid) and transition (dashed) macro F1 (\%) across horizons for best models per modality. Shaded regions show cross-subject standard deviation for the best (MViT-DCL (C)) and worst (DCL (a/g)).}
\label{fig:horizon_plot}
\end{minipage}
\end{table*}

Table~\ref{tab:locomotion_mode_recognition_results} presents results for current-state ($\tau{=}0$\,s) and predictive ($\tau{=}0.5$\,s) locomotion mode recognition. Within this standardized environment, the best visual models substantially outperform the inertial-only baselines. The per-class breakdown in Table~\ref{tab:per_class_f1} reveals complementary failure modes at $\tau{=}0$\,s: ResNet performs strongly on most terrain-driven classes but struggles on visually similar actions such as sit-to-stand ($51.6\%$) and stand-to-sit ($46.2\%$), where the temporal dynamics in the inertial signal give DCL an advantage. MViT improves upon ResNet on these temporally defined classes, highlighting the value of motion cues beyond static scene appearance. Per-class scores degrade at longer horizons, particularly during transitions; additional breakdowns are in Appendix~D.

Inertial--visual fusion yields the strongest overall performance by combining complementary kinematic and scene information. Across inertial+image methods, simple fusion already improves substantially over either modality alone, and the gains are most pronounced during transitions (e.g., the best inertial+image transition F1 of $66.2\%$ at $\tau{=}0$\,s vs.\ $53.4\%$ inertial-only and $54.2\%$ image-only). The best inertial+video model, MViT-DCL, reaches an overall F1 of $92.8\%$ at $\tau{=}0$\,s, dropping to $89.6\%$ at $\tau{=}0.5$\,s. Figure~\ref{fig:horizon_plot} shows that inertial-only performance degrades most steeply with increasing prediction horizon, while fusion models retain the highest scores and degrade more gradually. Transition performance, however, remains substantially below overall performance ($68.2\%$ vs.\ $92.8\%$ at $\tau{=}0$\,s), underscoring that accurately classifying task boundaries is still a difficult open problem.

\subsection{Cross-Population Generalization}

\begin{table*}[!t]
\centering
\begin{minipage}[t]{0.51\textwidth}
\vspace*{0pt}
\centering
\scriptsize
\setlength{\tabcolsep}{1.5pt}
\renewcommand{\arraystretch}{1.06}
\begin{tabular}{lcccc}
\toprule
\textbf{Method} & \multicolumn{2}{c}{$\boldsymbol{\tau{=}0.0}$\,s} & \multicolumn{2}{c}{$\boldsymbol{\tau{=}0.5}$\,s} \\
\cmidrule(lr){2-3} \cmidrule(lr){4-5}
 & Overall & Trans. & Overall & Trans. \\
\midrule
\rowcolor{modHeader}\multicolumn{5}{l}{\textit{\textbf{Inertial}}} \\
DCL (a)         & 59.9$\pm$10.3 & 34.7$\pm$10.7 & 53.5$\pm$9.7  & 33.5$\pm$8.9 \\
DCL (a/g)       & 65.6$\pm$9.5  & 36.8$\pm$11.8 & 59.3$\pm$8.7  & 37.9$\pm$9.9 \\
\midrule
\rowcolor{modHeader}\multicolumn{5}{l}{\textit{\textbf{Image}}} \\
MNV3-S          & 81.9$\pm$2.5  & 54.3$\pm$2.1  & 79.5$\pm$3.1  & 50.5$\pm$1.0 \\
R18             & 82.0$\pm$2.5  & 52.6$\pm$1.1  & 80.6$\pm$2.8  & 51.2$\pm$1.3 \\
R50             & 81.8$\pm$2.6  & 52.5$\pm$2.2  & 80.1$\pm$2.1  & 50.5$\pm$1.5 \\
\midrule
\rowcolor{modHeader}\multicolumn{5}{l}{\textit{\textbf{Video}}} \\
X3D-XS          & 80.5$\pm$2.7  & 46.5$\pm$4.5  & 81.8$\pm$1.5  & 39.8$\pm$6.0 \\
MViT            & 86.7$\pm$2.8  & \textbf{57.6$\pm$3.6} & 84.6$\pm$3.1 & \textbf{54.3$\pm$3.5} \\
\midrule
\rowcolor{modHeader}\multicolumn{5}{l}{\textit{\textbf{Inertial + Image}}} \\
KIFNet-Style    & 68.6$\pm$7.3  & 38.3$\pm$8.4  & 67.9$\pm$7.9  & 42.6$\pm$8.4 \\
MNV3-DCL (C)    & 87.1$\pm$1.9  & 56.5$\pm$6.1  & 83.5$\pm$1.2  & 52.3$\pm$3.6 \\
R18-DCL (C)     & 87.1$\pm$2.5  & 54.3$\pm$7.6  & 83.6$\pm$1.1  & 50.2$\pm$5.3 \\
\midrule
\rowcolor{modHeader}\multicolumn{5}{l}{\textit{\textbf{Inertial + Video}}} \\
SFTIK           & 84.0$\pm$3.1  & 52.3$\pm$3.8  & 82.1$\pm$3.6  & 53.9$\pm$2.6 \\
IMU-Video-MAE   & 86.5$\pm$2.8  & 55.4$\pm$6.1  & 85.2$\pm$2.3  & 53.8$\pm$3.8 \\
X3D-XS-DCL (C)  & 85.4$\pm$3.7  & 43.9$\pm$9.8  & 76.2$\pm$3.3  & 31.3$\pm$8.3 \\
MViT-DCL (C)    & \textbf{88.1$\pm$2.4} & 56.1$\pm$5.7 & \textbf{85.4$\pm$1.4} & 53.5$\pm$3.7 \\
\midrule
\midrule
\rowcolor{modHeader}\multicolumn{5}{l}{\textit{\textbf{Frozen Visual Backbone}}} \\
R18        & 73.7$\pm$3.3  & 49.2$\pm$2.2  & 74.5$\pm$3.4  & 47.4$\pm$2.9 \\
MViT       & 75.4$\pm$4.4  & 49.2$\pm$3.9  & 76.0$\pm$4.1  & 47.4$\pm$2.6 \\
R18-DCL    & 79.8$\pm$5.0  & 48.9$\pm$11.7 & 76.6$\pm$3.4  & 44.3$\pm$8.9 \\
\midrule
\midrule
\rowcolor{modHeader}\multicolumn{5}{l}{\textit{\textbf{Full Clinical Cohorts (Inertial Only), DCL (a/g)}}} \\
ST ($N{=}10$) & 57.7$\pm$19.0 & 35.0$\pm$13.1 & 51.2$\pm$17.1 & 32.9$\pm$11.1 \\
SR ($N{=}10$) & 39.9$\pm$20.7 & 22.1$\pm$9.8  & 34.8$\pm$19.3 & 21.5$\pm$11.5 \\
\bottomrule
\end{tabular}

\caption{Cross-population generalization at $\tau{=}0.0$\,s and $\tau{=}0.5$\,s, training on 7 non-impaired older adults. Mean $\pm$ SD macro F1 (\%) over 6 multimodal stroke survivors for all rows except the final block, which reports inertial-only DCL on the full ST and SR cohorts ($N{=}10$ each). \textbf{Bold}: best per column. Frozen-backbone rows freeze the listed visual backbone; remaining components are trained. Abbreviations as in Tables~\ref{tab:locomotion_mode_recognition_results} and~\ref{tab:cross_modality_results}.}

\label{tab:cross_population_results}
\end{minipage}\hfill
\begin{minipage}[t]{0.47\textwidth}
\vspace{0pt}
\centering
\scriptsize
\setlength{\tabcolsep}{1.2pt}
\renewcommand{\arraystretch}{1.06}
\begin{tabular}{|l|cccc|cccc|}
\hline
 & \multicolumn{4}{c|}{\textbf{HC$\rightarrow$SR} ($N{=}10$)} & \multicolumn{4}{c|}{\textbf{HC$\rightarrow$ST} ($N{=}10$)} \\
\cline{2-9}
 & \multicolumn{2}{c|}{$\boldsymbol{\tau{=}0.0}$\textbf{s}} & \multicolumn{2}{c|}{$\boldsymbol{\tau{=}0.5}$\textbf{s}} & \multicolumn{2}{c|}{$\boldsymbol{\tau{=}0.0}$\textbf{s}} & \multicolumn{2}{c|}{$\boldsymbol{\tau{=}0.5}$\textbf{s}} \\
\cline{2-9}
\textbf{Method} & O. & T. & O. & T. & O. & T. & O. & T. \\
\hline
Baseline     & 39.9                & 22.1                & 34.8          & 21.5          & 57.7                & 35.0          & 51.2                & 32.9 \\
NKD          & 41.4                & 22.4                & 35.7          & 21.5          & 59.2                & 34.8          & 51.9                & 32.7 \\
KD           & 41.5                & 23.2                & 36.3          & \textbf{22.4} & 58.0                & 35.1          & 51.8                & 32.8 \\
FitNets      & 41.8                & 23.7                & 35.9          & 21.5          & 60.5$^{*}$          & 36.3$^{*}$    & 53.9$^{*}$          & \textbf{34.0}$^{*}$ \\
CRD          & 42.3                & 24.3$^{*}$          & 35.2          & 21.7          & 60.3$^{*}$          & 36.5          & 52.6                & 32.9 \\
CP           & 46.4$^{*}$          & 25.1$^{*}$          & 38.1          & 22.1          & 63.6$^{*}$          & 35.5          & 54.5$^{*}$          & 32.4 \\
\hline
CP+FitNets   & \textbf{47.3}$^{*}$ & \textbf{25.3}       & \textbf{38.5} & 21.5          & \textbf{64.6}$^{*}$ & \textbf{36.7} & \textbf{55.2}$^{*}$ & 32.3 \\
\hline
\end{tabular}
\caption{Vision-guided knowledge transfer at $\tau{=}0.0$\,s and $\tau{=}0.5$\,s. Mean overall (O) and transition (T) macro F1 (\%) across test subjects and 5 seeds. $^{*}$: paired Wilcoxon $p<0.05$ vs Baseline, the inertial-only DCL (a/g) trained without vision guidance. \textbf{Bold}: best per column. HC: non-impaired older adults; ST: stroke survivors; SR: sarcopenic older adults.}
\label{tab:cross_modality_results}

\vspace{3em}

\makeatletter\def\@captype{figure}\makeatother
\includegraphics[width=\linewidth]{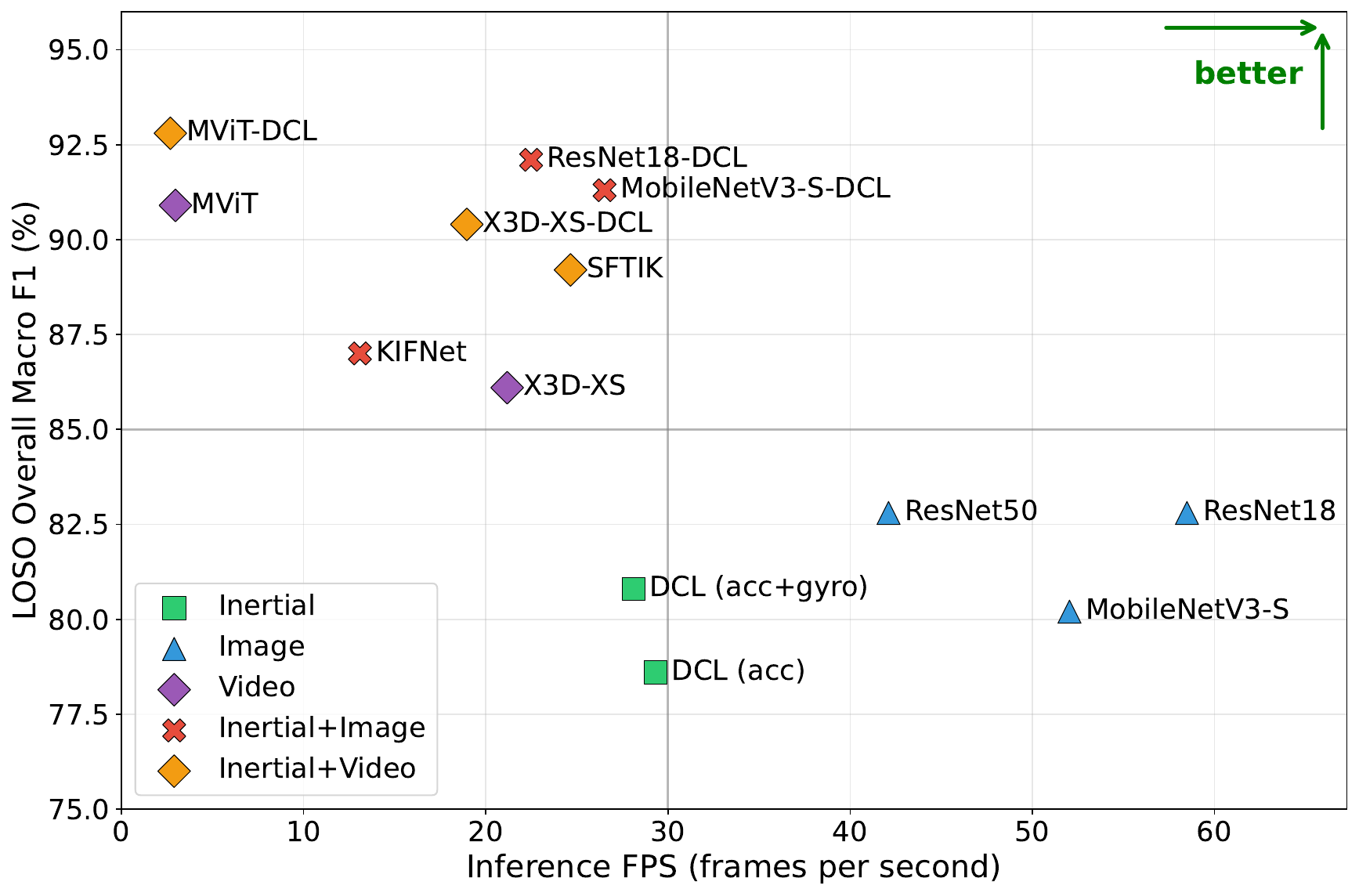}
\caption{Trade-off between overall LOSO macro F1 (\%) at $\tau{=}0$\,s and inference throughput (FPS) on a Jetson Orin Nano (8\,GB, 15\,W).}
\label{fig:jetson_latency}
\end{minipage}
\end{table*}

Table~\ref{tab:cross_population_results} shows that inertial-only models degrade most under cross-population evaluation and exhibit the highest inter-subject variability, reflecting cohort-level differences in how the tasks are performed (Figure~\ref{fig:task-statistics}): on the full clinical cohorts, overall F1 at $\tau{=}0$\,s reaches only $57.7\%$ for stroke survivors, while the sarcopenic cohort reaches only $39.9\%$. On the multimodal subset, vision transfers more consistently, with tighter dispersion across subjects, indicating that terrain and context cues remain useful despite differing gait kinematics, although this stability may also reflect the fixed scene layout. Multimodal fusion yields the strongest performance: the best fused model (MViT-DCL) reaches $88.1\%$ overall F1 at $\tau{=}0$\,s, dropping to $85.4\%$ at $\tau{=}0.5$\,s. Critically, both overall and transition scores are lower than in the LOSO setting (Table~\ref{tab:locomotion_mode_recognition_results}).

To test whether the visual gains require environment-specific backbone fine-tuning, the frozen-backbone block in Table~\ref{tab:cross_population_results} freezes the visual backbones while training the remaining task-specific components. Frozen ResNet-18 ($73.7\%$), MViT ($75.4\%$), and R18-DCL ($79.8\%$) still outperform the best inertial baseline ($65.6\%$) at $\tau{=}0$\,s, indicating that generic pretrained features provide useful terrain/context cues without environment-specific tuning. However, establishing cross-environment robustness still requires multi-site evaluation.

\subsection{Vision-Guided Knowledge Transfer}

Table~\ref{tab:cross_modality_results} evaluates vision-guided transfer to IMU-only inference on both clinical cohorts. Results are averaged over five training seeds per subject, followed by a two-sided paired Wilcoxon signed-rank test against the IMU-only baseline. CP is the strongest standalone method, significantly improving overall macro F1 in both cohorts at $\tau{=}0$\,s. KD and NKD yield no significant gains; CRD improves transition F1 for SR and overall F1 for ST, while FitNets significantly improves all four ST metrics. Initialized with CP weights, CP+FitNets achieves the best overall macro F1 at both horizons in both cohorts, exceeding the $\tau{=}0$\,s baseline by 7.4~pp on SR and 6.9~pp on ST. These results suggest that representation alignment and feature distillation are complementary, although transition recognition remains difficult and not all improvements reach significance at $N{=}10$.

\subsection{Inference Performance}
\label{sec:deployment}

Figure~\ref{fig:jetson_latency} summarizes end-to-end inference throughput (FPS) vs. accuracy on a Jetson Orin Nano (8\,GB, 15\,W). Single-frame RGB models offer the best throughput but underperform the top multimodal methods, while the most accurate models rely on heavier computations that substantially reduce throughput. Fusion models strike a practical balance, providing strong accuracy gains while remaining closer to deployability than large video transformers. While inference speed could improve further with TensorRT optimizations~\cite{10160419tsepa}, these results highlight an open challenge for assistive robotics: delivering transition-robust recognition under the compute, power, and latency constraints of wearable devices, motivating future work on efficient architectures that maintain robust performance within embedded budgets.

\section{Limitations and Future Work}
RevalExo's scope reflects design choices that prioritize clinically meaningful evaluation, and these choices open concrete opportunities for future work. First, it evaluates cross-population generalization in a standardized rehabilitation environment representative of clinical and training facilities increasingly used to evaluate assistive devices. This strengthens comparisons across cohorts under the same protocol, but means that visual models may exploit the fixed scene layout. The frozen-backbone control in Table~\ref{tab:cross_population_results} indicates that generic pretrained features provide useful visual context, but multi-site evaluation is needed to distinguish generalizable cues from fixed-scene effects and establish cross-environment robustness.

Second, the multimodal subset is shaped by clinical safety. As described in Section~\ref{sec:protocol}, the chest-mounted setup was omitted where its load introduced safety risks or reduced usable session time under the fatigue-limited protocol. The vision-guided knowledge transfer benchmark addresses this gap, supporting IMU-only deployment for cohorts without paired video data. A natural extension is lightweight, wireless capture hardware, which would extend multimodal coverage and strengthen future cross-population comparisons.

Third, we report transition performance as macro F1 within $\pm$0.25\,s windows around mode boundaries, capturing accuracy near transitions. Appendix~C complements this with event-level metrics (detection delay, missed transition rate, and false transition rate). Future work can build on this with lead-time evaluation~\cite{MarcosMazon2022,DalPreteTransition} and methods that balance transition detection speed and stability for real-time assistive control. Together, these benchmarks highlight two methodological gaps central to clinical assistive deployment: transition-window recognition remains far below overall recognition (e.g., $68.2\%$ vs.\ $92.8\%$ macro F1 for the best multimodal model), and inertial-only performance degrades substantially under cross-population transfer. RevalExo is designed to support method development on both fronts. Beyond locomotion mode recognition itself, its synchronized inertial and visual signals, frame-level annotations, and clinical cohort diversity also support complementary research directions, including temporal action segmentation~\cite{TemporalActionSegmentationSurvey} and self-supervised pretraining of multimodal motion-scene representations for downstream applications~\cite{CAREPD,GAITGEN}.

\section{Conclusion}
We introduced RevalExo, a functional daily-activity benchmark for inertial and visual locomotion mode recognition in older adults and clinical cohorts. RevalExo contains 10.1 hours of data with frame-level annotations from 27 participants across three cohorts: older adults without mobility impairments, stroke survivors, and older adults with probable sarcopenia. It includes 5.1 hours of synchronized egocentric video alongside the lower-body IMU recordings where clinically feasible. Through three benchmark tasks (locomotion mode recognition, cross-population generalization, and vision-guided knowledge transfer), we show that multimodal fusion provides the strongest overall performance, visual features retain higher performance than inertial features under population shift in this setup, and vision-guided training can improve IMU-only inference for cohorts without paired video. At the same time, RevalExo exposes persistent gaps that remain central for clinical assistive deployment: transition-window recognition remains far below overall recognition, and inertial-only performance degrades substantially under cross-population transfer. By releasing the dataset, annotations, and the benchmark codebase, we aim to foster further progress in assistive technologies that provide reliable, proactive support for individuals with mobility impairments.

\section*{Acknowledgements}
We thank all participants who gave their time and effort to take part in the study, and we thank Iulia Bilan, Mariam Akhalaia, and Lukas Varhol for their meticulous work on data annotation. This work was funded, in part, by the strategic basic research project RevalExo (S001024N) funded by the Research Foundation~--~Flanders (FWO), the AidWear project of the Federal Public Service for Policy and Support, and the Flemish Government under the Flanders AI Research Program (FAIR). Computational resources and services were provided by the Flemish Supercomputer Center (VSC), funded by the FWO and the Flemish Government.

\bibliography{main}

\clearpage
\appendix
\makeatletter
\let\arxiv@originalseccntformat\@seccntformat
\renewcommand{\@seccntformat}[1]{%
  \ifcsname arxiv@#1@prefix\endcsname
    \csname arxiv@#1@prefix\endcsname
  \else
    \arxiv@originalseccntformat{#1}%
  \fi
}
\newcommand{\arxiv@section@prefix}{\appendixname~\thesection\quad}
\makeatother

\section{Window Sampling}
\label{app:sampling}

Clips are $D{=}12$\,s long and the input window size is $W{=}2$\,s. Random sampling is seeded per epoch, so each epoch draws different windows while remaining reproducible across runs.

IMU data are recorded at 60\,Hz and egocentric video at 30\,fps. Within each 2\,s window, the IMU stream yields 120 time steps, while the video stream yields 60 frames. For video models, $T$ frames are uniformly subsampled from the 2\,s window. Each modality is encoded independently, and features are fused at the representation level rather than at the input level.

\begin{algorithm}[h]
\caption{Window sampling from a single clip}
\label{alg:sampling}
\begin{algorithmic}
\footnotesize
\Statex \textbf{Training} ($M{=}10$ windows per clip):
\State Divide the clip into $M$ equal temporal segments
\For{each segment}
    \State Randomly sample a start time $t$ within the segment
    \State Extract the window $[t,\; t + W]$
\EndFor
\Statex
\Statex \textbf{Evaluation} (stride $\delta{=}0.25$\,s):
\For{$t = 0,\; \delta,\; 2\delta,\; \dots$ until the end of the clip}
    \State Extract the window $[t,\; t + W]$
\EndFor
\end{algorithmic}
\end{algorithm}

\section{Effect of Input Window Size}
\label{app:window_size}

\begin{table}[ht]
\begin{center}
\scriptsize
\setlength{\tabcolsep}{5pt}
\renewcommand{\arraystretch}{1.08}
\begin{tabular}{|c|c|c|c|}
\hline
\textbf{Window (s)} & \textbf{$\tau{=}0.0$\,s} & \textbf{$\tau{=}0.5$\,s} & \textbf{FPS} \\
\hline
0.5 & 73.1 $\pm$ 8.2 & 65.9 $\pm$ 8.1 & 49.0 \\
1.0 & 76.3 $\pm$ 8.0 & 69.3 $\pm$ 7.9 & 36.6 \\
2.0 & 80.8 $\pm$ 8.0 & 74.2 $\pm$ 8.1 & 25.9 \\
3.0 & 81.5 $\pm$ 8.9 & 76.1 $\pm$ 8.9 & 14.4 \\
\hline
\end{tabular}
\end{center}
\caption[Effect of input window duration]{Effect of input window duration on DCL (a/g) recognition performance and inference speed on a Jetson Orin Nano. Macro F1 (\%) is reported as LOSO mean $\pm$ SD across subjects ($N=13$).}
\label{tab:window_size}
\end{table}

Table~\ref{tab:window_size} compares DCL (a/g) performance across input window durations from 0.5\,s to 3.0\,s. Increasing the window from 0.5\,s to 2.0\,s yields consistent gains at both horizons (+7.7~pp at $\tau{=}0$\,s, +8.3~pp at $\tau{=}0.5$\,s), while extending to 3.0\,s provides diminishing returns (+0.7~pp at $\tau{=}0$\,s, +1.9~pp at $\tau{=}0.5$\,s) at nearly half the inference speed.

\section{Transition Event Evaluation}
\label{app:transition_events}

We additionally evaluated dense 30\,Hz frame-level predictions (on the 13-subject multimodal subset) using a transition event protocol in line with prior locomotion mode recognition literature~\cite{10682545}. In Table~\ref{tab:transition_event}, we report: Detection delay (DD), the time from a ground-truth (GT) boundary to the first prediction of the new class; Missed transition rate (MTR@250\,ms), the fraction of transitions not detected within 250\,ms; False transition rate (FTR), predicted mode changes per minute not matching any GT transition; and Det., the fraction of transitions eventually detected before the next GT transition. Results are also reported after applying a 5-frame majority filter.

\begin{table}[t]
\begin{center}
\scriptsize
\setlength{\tabcolsep}{3pt}
\renewcommand{\arraystretch}{1.08}
\begin{tabular}{|l|c|c|c|c|c|}
\hline
\textbf{Model} & \textbf{MTR@250\,$\downarrow$} & \textbf{DD med.\,$\downarrow$} & \textbf{DD mean\,$\downarrow$} & \textbf{FTR/min\,$\downarrow$} & \textbf{Det.\,$\uparrow$} \\
\hline
\multicolumn{6}{|l|}{\textit{\textbf{Raw predictions}}} \\
\hline
DCL (IMU) & 24.5\% & 33\,ms  & 203\,ms & 33.6  & 95.1\% \\
ResNet-18 & 25.7\% & 0\,ms   & 92\,ms  & 122.1 & 85.1\% \\
Fusion    & 13.9\% & 21\,ms  & 70\,ms  & 55.0  & 95.5\% \\
\hline
\multicolumn{6}{|l|}{\textit{\textbf{5-frame majority filter}}} \\
\hline
DCL (IMU) & 32.9\% & 115\,ms & 288\,ms & 18.2 & 93.0\% \\
ResNet-18 & 39.4\% & 0\,ms   & 161\,ms & 21.3 & 76.9\% \\
Fusion    & 24.7\% & 49\,ms  & 146\,ms & 13.9 & 92.8\% \\
\hline
\end{tabular}
\end{center}
\caption{Event-level transition evaluation on dense 30\,Hz predictions. MTR@250: missed transition rate within 250\,ms ($\downarrow$ better). DD: detection delay ($\downarrow$ better). FTR: false transitions per minute ($\downarrow$ better). Det.: detection rate ($\uparrow$ better).}
\label{tab:transition_event}
\end{table}

Without filtering, fusion detects transitions in 70\,ms on average and misses only 13.9\% within 250\,ms, the best on both metrics; it also achieves the highest detection rate (95.5\%). The IMU-only DCL is more conservative, with a lower raw FTR (33.6 vs.\ 55.0) but slower detection (DD mean 203\,ms) and a higher miss rate (24.5\%). This delay is intrinsic to inertial-only sensing: DCL has no visual access to upcoming terrain, so for terrain-driven transitions (e.g., entering stairs, ramps, grass, or uneven ground) it cannot commit to the new mode until the gait pattern itself has changed. This only happens after the transition has already begun. ResNet-18 attains a DD median of 0\,ms because the image branch can react to a new terrain type the instant it enters the field of view, but it pays for this with the highest raw FTR (122.1) and the lowest detection rate (85.1\%); the 0\,ms median does not imply the best overall transition performance, since DD is computed only over detected transitions. The 5-frame majority filter reduces false transitions at the cost of higher latency, consistent with the stability-latency trade-off reported by Ma et al.~\cite{10682545}; for fusion this trade is particularly favourable, taking FTR from 55.0 to 13.9 per minute while detection rate only drops from 95.5\% to 92.8\%. Overall, fusion offers the best balance of detection rate, latency, and stability.

\section{Additional Per-Class Results}
\label{app:full_per_class_results}

\begin{table*}[!t]
\centering
\tiny
\setlength{\tabcolsep}{1.2pt}
\renewcommand{\arraystretch}{1.04}
\begin{minipage}[t]{0.325\textwidth}
\centering
\textbf{Transition, $\tau{=}0.0$\,s}\par\vspace{1mm}
\resizebox{\textwidth}{!}{%
\begin{tabular}{|l|*{5}{>{\centering\arraybackslash}m{3.1em}}|}
\hline
\diagbox[width=6.7em, height=3.5em, innerleftsep=2pt, innerrightsep=2pt]{\textbf{Class}}{\textbf{Method}}
 & \rotatebox[origin=lB]{68}{DCL (a/g)}
 & \rotatebox[origin=lB]{68}{R50}
 & \rotatebox[origin=lB]{68}{MViT}
 & \rotatebox[origin=lB]{68}{R18-DCL (C)}
 & \rotatebox[origin=lB]{68}{MViT-DCL (C)} \\
\noalign{\vspace*{-0em}}
\hline
Level ground  & 61.1 & \underline{50.2} & 56.5 & 63.0 & \textbf{64.7} \\
Sit to stand  & 67.4 & \underline{44.8} & 66.4 & 71.1 & \textbf{71.3} \\
Stand to sit  & 62.7 & \underline{34.6} & 62.4 & 69.9 & \textbf{72.3} \\
Sitting       & 61.6 & \underline{52.9} & 59.9 & 62.6 & \textbf{63.2} \\
Stair up      & 60.3 & \underline{60.0} & 67.6 & 72.0 & \textbf{75.7} \\
Stair down    & \underline{55.1} & 55.8 & 65.8 & \textbf{73.4} & 66.7 \\
Ramp up       & \underline{43.4} & 62.5 & 60.7 & 62.2 & \textbf{67.4} \\
Ramp down     & \underline{54.9} & 56.6 & 60.4 & 65.0 & \textbf{68.1} \\
Grass         & \underline{41.1} & 61.4 & 67.1 & 67.4 & \textbf{70.1} \\
Uneven ground & \underline{47.9} & 57.9 & 63.1 & 64.3 & \textbf{70.5} \\
Carry         & \underline{31.4} & 59.1 & \textbf{62.1} & 57.0 & 60.0 \\
\hline
\end{tabular}%
}
\end{minipage}
\hfill
\begin{minipage}[t]{0.325\textwidth}
\centering
\textbf{Overall, $\tau{=}0.5$\,s}\par\vspace{1mm}
\resizebox{\textwidth}{!}{%
\begin{tabular}{|l|*{5}{>{\centering\arraybackslash}m{3.1em}}|}
\hline
\diagbox[width=6.7em, height=3.5em, innerleftsep=2pt, innerrightsep=2pt]{\textbf{Class}}{\textbf{Method}}
 & \rotatebox[origin=lB]{68}{DCL (a/g)}
 & \rotatebox[origin=lB]{68}{R50}
 & \rotatebox[origin=lB]{68}{MViT}
 & \rotatebox[origin=lB]{68}{R18-DCL (C)}
 & \rotatebox[origin=lB]{68}{MViT-DCL (C)} \\
\noalign{\vspace*{-0em}}
\hline
Level ground  & \underline{75.6} & 86.2 & 89.6 & 89.6 & \textbf{90.8} \\
Sit to stand  & 69.5 & \underline{57.3} & 68.5 & 73.1 & \textbf{74.4} \\
Stand to sit  & 64.2 & \underline{55.6} & 70.5 & 71.9 & \textbf{76.7} \\
Sitting       & 83.6 & \underline{75.3} & 83.0 & \textbf{85.1} & 84.9 \\
Stair up      & \underline{77.3} & 89.9 & 90.9 & 91.2 & \textbf{91.6} \\
Stair down    & \underline{74.7} & 88.6 & 93.1 & 91.9 & \textbf{94.0} \\
Ramp up       & \underline{79.6} & 93.3 & 93.3 & 93.9 & \textbf{94.0} \\
Ramp down     & \underline{82.8} & 93.0 & 92.5 & \textbf{94.1} & 93.5 \\
Grass         & \underline{69.1} & 94.1 & 94.3 & 94.9 & \textbf{95.1} \\
Uneven ground & \underline{76.1} & 94.0 & 94.7 & 94.8 & \textbf{95.5} \\
Carry         & \underline{64.4} & 92.7 & 95.2 & 94.1 & \textbf{95.6} \\
\hline
\end{tabular}%
}
\end{minipage}
\hfill
\begin{minipage}[t]{0.325\textwidth}
\centering
\textbf{Transition, $\tau{=}0.5$\,s}\par\vspace{1mm}
\resizebox{\textwidth}{!}{%
\begin{tabular}{|l|*{5}{>{\centering\arraybackslash}m{3.1em}}|}
\hline
\diagbox[width=6.7em, height=3.5em, innerleftsep=2pt, innerrightsep=2pt]{\textbf{Class}}{\textbf{Method}}
 & \rotatebox[origin=lB]{68}{DCL (a/g)}
 & \rotatebox[origin=lB]{68}{R50}
 & \rotatebox[origin=lB]{68}{MViT}
 & \rotatebox[origin=lB]{68}{R18-DCL (C)}
 & \rotatebox[origin=lB]{68}{MViT-DCL (C)} \\
\noalign{\vspace*{-0em}}
\hline
Level ground  & 53.5 & \underline{49.7} & 54.4 & \textbf{56.8} & 56.0 \\
Sit to stand  & 59.3 & \underline{49.9} & 56.8 & 59.6 & \textbf{62.7} \\
Stand to sit  & 54.5 & \underline{38.7} & 50.3 & 52.5 & \textbf{58.2} \\
Sitting       & 54.2 & \underline{36.3} & 50.2 & \textbf{56.6} & 54.8 \\
Stair up      & \underline{48.7} & 65.5 & 62.2 & \textbf{68.5} & 67.5 \\
Stair down    & \underline{43.1} & 52.1 & 59.5 & 62.2 & \textbf{65.6} \\
Ramp up       & \underline{43.6} & 61.2 & 61.7 & 58.3 & \textbf{63.0} \\
Ramp down     & 54.4 & \underline{54.2} & 59.4 & 62.3 & \textbf{64.5} \\
Grass         & \underline{41.1} & 59.8 & \textbf{63.7} & 61.9 & 63.2 \\
Uneven ground & \underline{46.1} & 54.9 & 61.0 & 58.4 & \textbf{66.7} \\
Carry         & \underline{25.8} & 55.2 & \textbf{59.3} & 55.2 & 58.5 \\
\hline
\end{tabular}%
}
\end{minipage}
\caption{Additional subject-averaged per-class F1 (\%) for the top-performing methods per modality. The main paper reports overall scores at $\tau{=}0.0$\,s; these tables report transition-window scores at $\tau{=}0.0$\,s, overall scores at $\tau{=}0.5$\,s, and transition-window scores at $\tau{=}0.5$\,s. \textbf{Bold}: best per class. \underline{Underline}: worst per class.}
\label{tab:per_class_appendix}
\end{table*}

Table~\ref{tab:per_class_appendix} provides the remaining per-class breakdowns for the same representative models shown in the main paper. The additional results show that per-class degradation at longer horizons is most pronounced during transition windows, where short and boundary-sensitive classes are harder to classify reliably.

\section{Full Per-Horizon Results}
\label{app:full_results}

\begin{table*}[t]
\centering
\scriptsize
\setlength{\tabcolsep}{1.2pt}
\renewcommand{\arraystretch}{1.04}
\resizebox{\textwidth}{!}{%
\begin{tabular}{lcccccccccccc}
\toprule
\textbf{Method} & \multicolumn{2}{c}{$\boldsymbol{\tau{=}0.0}$\,s} & \multicolumn{2}{c}{$\boldsymbol{\tau{=}0.1}$\,s} & \multicolumn{2}{c}{$\boldsymbol{\tau{=}0.2}$\,s} & \multicolumn{2}{c}{$\boldsymbol{\tau{=}0.3}$\,s} & \multicolumn{2}{c}{$\boldsymbol{\tau{=}0.5}$\,s} & \multicolumn{2}{c}{$\boldsymbol{\tau{=}1.0}$\,s} \\
\cmidrule(lr){2-3} \cmidrule(lr){4-5} \cmidrule(lr){6-7} \cmidrule(lr){8-9} \cmidrule(lr){10-11} \cmidrule(lr){12-13}
 & Overall & Trans. & Overall & Trans. & Overall & Trans. & Overall & Trans. & Overall & Trans. & Overall & Trans. \\
\midrule
\rowcolor{modHeader}\multicolumn{13}{l}{\textit{\textbf{Inertial}}} \\
DCL (a)         & 78.6$\pm$8.6 & 50.6$\pm$9.8 & 76.9$\pm$9.0 & 49.3$\pm$10.2 & 75.7$\pm$9.2 & 49.9$\pm$10.1 & 74.1$\pm$9.4 & 48.4$\pm$9.2 & 70.9$\pm$9.2 & 46.4$\pm$8.1 & 63.1$\pm$8.9 & 40.3$\pm$8.8 \\
DCL (a/g)       & 80.8$\pm$8.0 & 53.4$\pm$10.1 & 79.6$\pm$8.1 & 52.2$\pm$9.8 & 78.5$\pm$8.2 & 50.6$\pm$9.8 & 77.1$\pm$8.1 & 50.3$\pm$8.6 & 74.2$\pm$8.1 & 47.7$\pm$8.3 & 66.9$\pm$7.9 & 43.0$\pm$5.7 \\
\midrule
\rowcolor{modHeader}\multicolumn{13}{l}{\textit{\textbf{Image}}} \\
MNV3-S          & 80.2$\pm$2.6 & 52.1$\pm$3.3 & 80.6$\pm$2.5 & 52.1$\pm$2.5 & 81.0$\pm$2.5 & 52.2$\pm$2.7 & 81.4$\pm$2.3 & 51.5$\pm$2.3 & 81.5$\pm$2.3 & 50.5$\pm$3.7 & 77.6$\pm$3.1 & 46.4$\pm$5.3 \\
R18             & 82.8$\pm$2.1 & 53.8$\pm$3.0 & 83.2$\pm$1.9 & 53.8$\pm$2.4 & 83.4$\pm$1.8 & 53.7$\pm$2.9 & 83.4$\pm$1.7 & 52.0$\pm$3.1 & 83.2$\pm$1.7 & 52.2$\pm$3.2 & 79.0$\pm$2.7 & 47.1$\pm$4.6 \\
R50             & 82.8$\pm$1.8 & 54.2$\pm$3.1 & 83.2$\pm$1.9 & 54.0$\pm$3.2 & 83.5$\pm$2.0 & 54.6$\pm$2.9 & 83.7$\pm$2.1 & 53.5$\pm$2.8 & 83.6$\pm$2.0 & 52.5$\pm$3.3 & 80.1$\pm$2.6 & 48.7$\pm$5.3 \\
\midrule
\rowcolor{modHeader}\multicolumn{13}{l}{\textit{\textbf{Video}}} \\
X3D-XS          & 86.1$\pm$3.0 & 47.8$\pm$6.9 & 85.9$\pm$3.1 & 47.2$\pm$8.4 & 85.2$\pm$3.6 & 44.5$\pm$9.2 & 84.4$\pm$3.9 & 42.5$\pm$9.0 & 82.4$\pm$4.6 & 40.0$\pm$7.9 & 77.2$\pm$5.2 & 38.4$\pm$6.9 \\
MViT            & 90.9$\pm$2.2 & 62.9$\pm$3.6 & 90.6$\pm$2.2 & 62.7$\pm$3.2 & 89.8$\pm$2.2 & 60.8$\pm$4.0 & 89.3$\pm$2.3 & 61.0$\pm$3.4 & 87.8$\pm$2.2 & 58.0$\pm$3.5 & 84.0$\pm$2.1 & 51.7$\pm$5.4 \\
\midrule
\rowcolor{modHeader}\multicolumn{13}{l}{\textit{\textbf{Inertial + Image}}} \\
KIFNet-Style    & 84.7$\pm$2.5 & 51.5$\pm$7.0 & 84.3$\pm$2.1 & 51.4$\pm$6.6 & 83.6$\pm$2.1 & 52.1$\pm$5.4 & 82.6$\pm$2.3 & 50.4$\pm$5.7 & 80.8$\pm$2.8 & 47.8$\pm$6.7 & 77.2$\pm$2.4 & 43.2$\pm$6.4 \\
KIFNet (A)      & 87.5$\pm$2.7 & 55.2$\pm$7.9 & 86.9$\pm$2.4 & 54.7$\pm$6.6 & 86.2$\pm$2.2 & 54.4$\pm$5.3 & 85.2$\pm$2.1 & 51.8$\pm$6.4 & 83.7$\pm$2.1 & 49.3$\pm$6.4 & 80.1$\pm$2.6 & 45.2$\pm$6.2 \\
KIFNet (C)      & 87.0$\pm$3.0 & 55.5$\pm$6.1 & 86.4$\pm$2.6 & 55.0$\pm$6.1 & 85.7$\pm$2.3 & 54.0$\pm$5.3 & 85.1$\pm$2.1 & 53.4$\pm$5.7 & 83.6$\pm$2.0 & 50.8$\pm$5.8 & 79.9$\pm$2.4 & 45.0$\pm$8.3 \\
MNV3-DCL (A)    & 91.4$\pm$1.7 & 65.1$\pm$6.1 & 90.9$\pm$1.8 & 63.9$\pm$5.5 & 90.4$\pm$2.0 & 62.7$\pm$4.9 & 89.6$\pm$2.1 & 60.3$\pm$6.2 & 88.1$\pm$2.2 & 57.8$\pm$5.6 & 83.6$\pm$2.2 & 52.1$\pm$3.8 \\
MNV3-DCL (C)    & 91.3$\pm$1.6 & 65.8$\pm$4.0 & 90.8$\pm$1.7 & 64.6$\pm$4.6 & 90.2$\pm$1.9 & 64.0$\pm$4.4 & 89.4$\pm$2.2 & 61.8$\pm$3.6 & 87.9$\pm$2.4 & 59.4$\pm$4.7 & 84.0$\pm$2.7 & 52.7$\pm$4.3 \\
R18-DCL (A)     & 92.3$\pm$1.7 & 65.9$\pm$5.3 & 91.8$\pm$1.7 & 65.2$\pm$5.0 & 91.2$\pm$1.8 & 63.6$\pm$5.5 & 90.4$\pm$1.9 & 61.7$\pm$4.6 & 88.5$\pm$2.0 & 58.7$\pm$4.2 & 84.0$\pm$2.0 & 52.7$\pm$4.4 \\
R18-DCL (C)     & 92.1$\pm$1.7 & 66.2$\pm$5.0 & 91.6$\pm$1.8 & 65.1$\pm$4.8 & 91.0$\pm$1.8 & 63.4$\pm$5.2 & 90.3$\pm$2.0 & 61.8$\pm$6.2 & 88.6$\pm$2.1 & 59.3$\pm$5.5 & 84.5$\pm$2.2 & 52.9$\pm$3.9 \\
\midrule
\rowcolor{modHeader}\multicolumn{13}{l}{\textit{\textbf{Inertial + Video}}} \\
SFTIK           & 89.2$\pm$3.5 & 59.8$\pm$4.8 & 88.8$\pm$3.5 & 60.1$\pm$4.0 & 88.4$\pm$3.5 & 60.3$\pm$4.9 & 87.9$\pm$3.2 & 59.7$\pm$4.3 & 86.9$\pm$2.9 & 58.0$\pm$4.4 & 83.8$\pm$3.2 & 52.2$\pm$4.5 \\
IMU-Video-MAE   & 90.3$\pm$2.0 & 61.8$\pm$4.8 & 90.1$\pm$2.1 & 61.3$\pm$4.7 & 89.7$\pm$2.2 & 60.4$\pm$6.3 & 89.2$\pm$2.2 & 59.6$\pm$5.4 & 88.0$\pm$2.4 & 57.5$\pm$5.8 & 83.9$\pm$2.4 & 52.2$\pm$4.8 \\
X3D-XS-DCL (A)  & 90.1$\pm$3.8 & 56.2$\pm$7.7 & 89.3$\pm$4.4 & 53.5$\pm$8.7 & 88.5$\pm$4.7 & 49.2$\pm$8.5 & 86.7$\pm$5.0 & 45.4$\pm$8.8 & 83.2$\pm$5.6 & 41.3$\pm$9.0 & 75.6$\pm$6.0 & 39.3$\pm$7.6 \\
X3D-XS-DCL (C)  & 91.1$\pm$2.9 & 60.2$\pm$6.9 & 90.6$\pm$3.2 & 59.1$\pm$5.8 & 89.9$\pm$3.4 & 54.7$\pm$7.8 & 88.7$\pm$3.7 & 51.5$\pm$7.8 & 85.7$\pm$4.0 & 47.3$\pm$7.1 & 78.5$\pm$4.2 & 42.7$\pm$5.2 \\
MViT-DCL (A)    & 92.7$\pm$1.7 & 67.6$\pm$4.6 & 92.3$\pm$2.1 & \textbf{67.1$\pm$4.1} & 91.7$\pm$2.2 & 65.0$\pm$5.3 & 90.8$\pm$2.3 & 63.8$\pm$4.8 & 88.8$\pm$2.5 & 61.0$\pm$5.2 & 85.1$\pm$2.7 & 54.0$\pm$4.6 \\
MViT-DCL (C)    & \textbf{92.8$\pm$2.0} & \textbf{68.2$\pm$5.4} & \textbf{92.4$\pm$2.1} & \textbf{67.1$\pm$4.4} & \textbf{92.0$\pm$2.1} & \textbf{65.9$\pm$5.3} & \textbf{91.3$\pm$2.2} & \textbf{64.6$\pm$4.6} & \textbf{89.6$\pm$2.3} & \textbf{61.9$\pm$4.5} & \textbf{85.8$\pm$2.2} & \textbf{56.0$\pm$3.7} \\
\bottomrule
\end{tabular}}
\caption{Full per-horizon locomotion mode recognition results. Mean $\pm$ SD macro F1 (\%) over the 13-subject multimodal subset (LOSO-CV). \textbf{Bold}: best per column. DCL: DeepConvLSTM; MNV3-S: MobileNet-v3 Small; R18/R50: ResNet-18/-50; a/g: accelerometer/gyroscope; A/C: feature averaging/concatenation.}
\label{tab:full_results}
\end{table*}

Table~\ref{tab:full_results} extends the main locomotion mode recognition table (Section 5.1) with results across all six prediction horizons. Training logs are available from \href{https://huggingface.co/datasets/wearablehar/train-logs/tree/main}{Hugging Face}.

\end{document}